\pdfoutput=1
\documentclass[11pt]{article}

\usepackage[final]{acl}

\usepackage{times}
\usepackage{latexsym}

\usepackage[T1]{fontenc}
\usepackage[utf8]{inputenc}

\usepackage{microtype}

\usepackage{inconsolata}

\usepackage{graphicx}

\usepackage{amsthm, algorithm, amsmath, amsfonts, mleftright, multirow, bm, booktabs}

\usepackage{colortbl}
\definecolor{lightgray}{rgb}{0.95, 0.95, 0.95}
\newcolumntype{g}{>{\columncolor{lightgray}}c}

\title{Initialization of Large Language Models via Reparameterization \\
to Mitigate Loss Spikes}
\author{
		Kosuke Nishida \quad\quad%\quad
		Kyosuke Nishida \quad\quad%\quad
		Kuniko Saito \\
        NTT Human Informatics Laboratories, NTT Corporation \\
		\tt \{kosuke.nishida, kyosuke.nishida, kuniko.saito\}@ntt.com
}

\begin{document}
\maketitle
\begin{abstract}
Loss spikes, a phenomenon in which the loss value diverges suddenly, is a fundamental issue in the pre-training of large language models. This paper supposes that the non-uniformity of the norm of the parameters is one of the causes of loss spikes. Here, in training of neural networks, the scale of the gradients is required to be kept constant throughout the layers to avoid the vanishing and exploding gradients problem. However, to meet these requirements in the Transformer model, the norm of the model parameters must be non-uniform, and thus, parameters whose norm is smaller are more sensitive to the parameter update. To address this issue, we propose a novel technique, weight scaling as reparameterization (WeSaR). WeSaR introduces a gate parameter per parameter matrix and adjusts it to the value satisfying the requirements. Because of the gate parameter, WeSaR sets the norm of the original parameters uniformly, which results in stable training. Experimental results with the Transformer decoders consisting of 130 million, 1.3 billion, and 13 billion parameters showed that WeSaR stabilizes and accelerates training and that it outperformed compared methods including popular initialization methods.
\end{abstract}

\section{Introduction}
\begin{figure}[t!]
\centering
\includegraphics[width=0.8\linewidth]{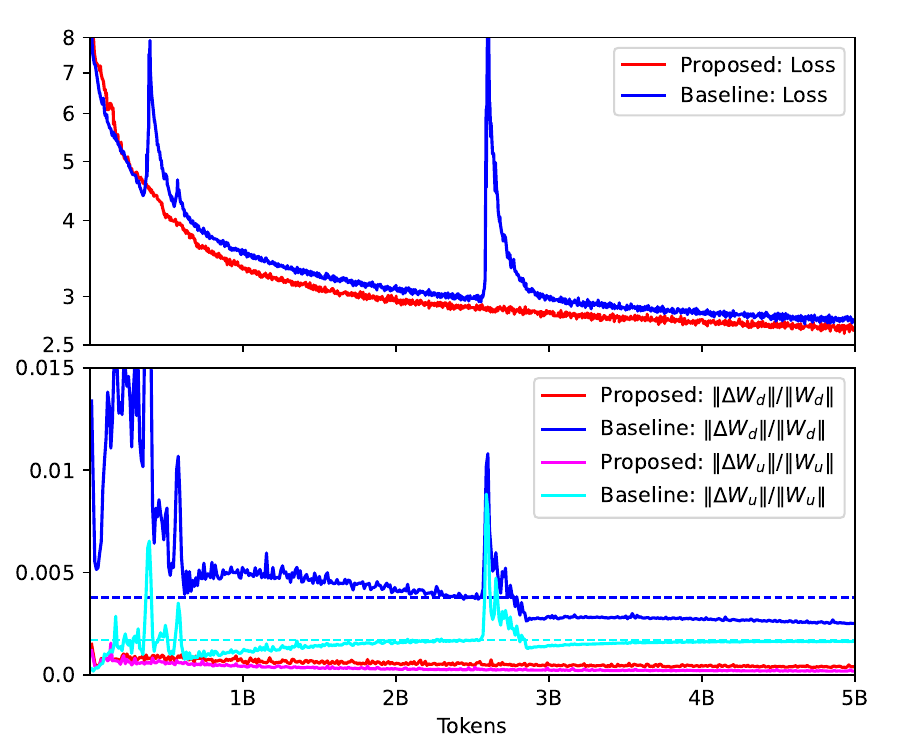}
		\caption{Loss of Transformer models with 13 billion (13B) parameters at the beginning of training (top). 
%Update ratios $\|\Delta \bm{W}_d\|/\| \bm{W}_d\|$ and $\|\Delta \bm{W}_u\|/\| \bm{W}_u\|$ of the same (bottom). 
Update ratios for the up and down projection in the last feed-forward layer, $\|\Delta \bm{W}_u\|/\| \bm{W}_u\|$ and $\|\Delta \bm{W}_d\|/\| \bm{W}_d\|$, of the same (bottom). 
The horizontal lines are the update ratios before the largest spike. The baseline sets $\| \bm{W}_d\|$ smaller than the other parameters. The update ratio of $\bm{W}_d$ %$\|\Delta \bm{W}_d\|/\| \bm{W}_d\|$ 
        is larger at the very beginning and gets smaller after loss spikes occur. The baseline uses standard techniques for stable training, such as gradient clipping.}
        \label{fig:13B_init_loss}
\end{figure}

Transformer-based large language models (LLMs) have attracted remarkable attention~\cite{transformer, gpt-3}. The discovery of a scaling-law~\cite{kaplan2020scaling} has been driving the model and corpus sizes ever larger, causing huge computational costs for pre-training.
During pre-training of LLMs, the loss value often diverges suddenly~\cite{palm, opt}, as illustrated at the top of Figure~\ref{fig:13B_init_loss}. This phenomenon, known as loss spikes, is a fundamental issue in the LLM pre-training because it not only increases the final loss value, but also causes the pre-training to fail if the loss diverges completely. 
%With respect to the model performance, it increases the final loss value. More important issue is regarding the pre-training cost. it fails the pre-training if the loss diverges completely.

%This paper indicates the non-uniformity of the norm of the parameters as one of the causes of loss spikes, as discussed in \S\ref{sec:problem}.
Here, let $\Delta \bm{W}$ be the update of the parameter $\bm{W}$ at an optimization step. $\|\Delta \bm{W}\| / \|\bm{W}\|$ represents the magnitude of the parameter update relative to the parameter itself, and we call it the \textit{update ratio}. %denote it by $\rho$. 
The bottom of Figure~\ref{fig:13B_init_loss} shows the update ratios. 
%明らかに異なる傾向〜について、updateratioは単にlossの値に比例するものではない、みたいな説明を明確に書いてみるのもどうかな。
%we observed thatのあたりですが、学習の最初期やspike直後などの不安定な時期では、updateratioにも大きな変動が出る。また、spikeの後のratioはspikeの前とは明らかに異なる傾向が見られる、lossが回復しているように見えてもモデル内部には影響が残っている
%\textcolor{blue}{We observe that the update ratio of $\bm{W}_d$ after the loss spike is smaller than that before the spike, as discussed in \S\ref{sec:problem}. Because an update ratio is the relative magnitude of the update, a large update ratio can lead to unstable training and loss spikes.} 
%That is, informally, if the update ratio is stable, the training is stable, and vice versa. However, as shown at the bottom of Figure~\ref{fig:13B_init_loss}, a baseline pre-training method makes the update ratio unstable. Moreover, the divergence of the update ratio causes loss spikes, and the update ratio becomes lower than those before the spike.
%\textcolor{red}{When using the baseline initialization method currently standard for LLM training, we observed instability in the update ratios during the early stages of training and at times of loss spikes. Notably, the update ratios for $\bm{W}_d$ changed around the time of these loss spikes, as indicated by the dotted line in Figure 1. This ratio does not simply correlate with loss values or learning rates; rather, it may serve as an indicator of training stability}. 
We consider that different scales of update ratios among parameter matrices can lead to unstable training. Indeed, before the loss spike, the update ratio of $\bm{W}_d$ is larger than that of $\bm{W}_u$. That is, $\bm{W}_d$ undergoes a more pronounced change.
After the spike, the difference between the update ratios decreases.
This observation motivated us to regulate the update ratios in the model in a certain range.
%Notably, the update ratio for the down projection in feed-forward networks, $||\Delta \bm{W}_d||/||\bm{W}_d||$, is smaller after a loss spike compared to before, despite similar loss values before and after the spike}.

%\textcolor{red}{We consider that such instability in the update ratios is caused by the non-uniformity of the parameter norms. With the current initialization methods, $\bm{W}_d$ is set smaller than other parameters to stabilize the gradients. Consequently, by definition, the update ratios for $\bm{W}_d$ tend to be larger, which can often lead to loss spikes}.

%\textcolor{blue}{We consider that this large update ratio is due to the non-uniformity of the norm of the parameters. That is, by definition, the update ratio is large in parameters whose norm is smaller than those of other parameters. In Figure~\ref{fig:13B_init_loss}, $\|\bm{W}_d\|$ is smaller than $\|\bm{W}_u\|$. However, the non-uniformity is inevitable because it arises from the initialization strategies of the Transformer that are designed to avoid the vanishing and exploding gradients problem, as discussed in \S\ref{sec:preliminary}.} 

We consider that uneven and large update ratios are due to non-uniformity of the norm of the parameters. With the current initialization methods, $\bm{W}_d$ is set smaller than other parameters, which is required to avoid the vanishing and exploding gradients problem. Consequently, by definition, the update ratio of $\bm{W}_d$ tends to be larger.%, which can lead to training instability. %However, the non-uniformity is inevitable for the Transformer architecture to avoid the vanishing and exploding gradients problem. %, as discussed in \S\ref{sec:preliminary}.

To address this issue, we propose a novel technique, called \underline{we}ight \underline{s}caling \underline{a}s \underline{r}eparameterization (WeSaR). WeSaR introduces a gate parameter $\alpha \in \mathbb{R}$ for each parameter matrix $\bm{W}$ and uses
%\textcolor{blue}{rearranges the Transformer model by introducing} a gate parameter $\alpha \in \mathbb{R}$ for each parameter matrix $W$ and using 
$\alpha \bm{W}$ instead of $\bm{W}$ inside the model. WeSaR relieves the parameter $\bm{W}$ of non-uniformity by adjusting $\alpha$ to the values required to avoid the vanishing and exploding gradients problem.
Moreover, WeSaR enables an arbitrary small common standard deviation to set be for all parameters, which results in not only stable, but also accelerated, training.
%そこで本研究では，新規にゲートパラメータ$\alpha \in \mathbb{R}$を導入し，各パラメータ行列$W$の代わりに，$\alpha W$をモデル内部で用いることを提案する．提案手法のGating Initializationは，勾配爆発・消失を回避するために要求される初期値の標準偏差に関する制約を，ゲートパラメータ$\alpha$を所定の値に指定することで満たすことを特徴とする．また，既存の深層学習の初期化手法と異なり，$W$の初期値の標準偏差を任意の値に設定可能とする．

We conducted pre-training of Transformer decoders consisting of 130 million (13M), 1.3 billion (1.3B), and 13B parameters. Our experimental results show that WeSaR stabilized and accelerated their training due to the stable and equal-scale update ratios, as shown in Figure~\ref{fig:13B_init_loss}. We also confirmed that WeSaR outperformed compared methods, including a initialization method widely used for pre-training LLMs~\cite{nguyen-salazar-2019-transformers} and the existing reparameterization methods~\cite{NIPS2016_ed265bc9,pmlr-v202-zhai23a,NEURIPS2022_ae0cba71}.
%評価実験により，Gating Initializationによって安定に学習が進み，損失関数値の小さいモデルが得られることを確認した．%さらに，LLMの事前学習は安定な学習を実現するために小さい学習率を採用するなど，安定であるが学習速度が遅いハイパーパラメータを採用している．提案手法によって，学習速度を重視したハイパーパラーメータを採用した場合でも安定してLLMを学習することができることを確認した．LLMの事前学習においては用意したデータ量によって性能が規定されるが，本研究によって同じデータで学習したLLMの性能を大きく引き上げることができる．

Our contributions can be summarized as follows:
\begin{itemize}
    \item We clarify one of the causes of loss spikes, \textit{i.e.,} the non-uniformity of parameters that arises to meet the requirements for avoiding the vanishing and exploding gradients problem. % induces the instability.
    \item We address the non-uniformity problem by reparameterizing the parameter as $\alpha \bm{W}$ with a gate parameter $\alpha$. $\alpha$ determines the scale of $\alpha \bm{W}$. $\bm{W}$ is initialized with a small common standard deviation throughout the model.
    \item Experimental results show that the proposed method stabilizes and accelerates training. It outperformed compared methods, including a popular initialization method of LLMs.
\end{itemize}

\section{Preliminaries}
\label{sec:preliminary}
We consider Transformer models~\cite{transformer} consisting of the following layers: an embedding layer with $\bm{W}_e$, self-attention layers with $\bm{W}_q$, $\bm{W}_k$, $\bm{W}_v$, and $\bm{W}_o$ (query, key, value, and output projections), feed-forward layers with $\bm{W}_u$ and $\bm{W}_d$ (up and down projections)\footnote{We did not use GLU~\cite{shazeer2020glu} for simplicity.}, and a prediction layer with $\bm{W}_p$.
Each parameter $\bm{W}_{\cdot}$ is initialized according to a Gaussian distribution $\mathcal{N}(0, \sigma_{\cdot}^2)$. %In this subsection, we discuss two strategies for determining the standard deviations to achieve the aforementioned demand.

The input first passes through the embedding layer; then it is processed by $N$ Transformer blocks, which consist of self-attention layers and feed-forward layers. The transformation $f$ of the self-attention layer and 
the feed-forward layer with a residual connection can be written as 
\begin{align}
\bm{y} =f(\mathrm{LN}(\bm{x})) + \bm{x},
\label{eq:residual_forward}
\end{align}
where LN indicates a layer normalization~\cite{layer_norm} that is applied after the residual connection, called the Pre-LN type~\cite{liu-etal-2020-understanding}.

%\textcolor{red}{We consider RMSNorm~\cite{rmsnorm}, which has been  widely used in modern LLMs such as Llama-3 instead of LayerNorm~\cite{layer_norm}, and Pre-LN, which is reported to stabilize the training~\cite{liu-etal-2020-understanding}}.
%Layer Normalization~\cite{layer_norm}.
%RMSNorm.

%To stabilize training, two strategies are employed: initialization and normalization. 
In this section, we first review the back-propagation algorithm% for the case of linear layers
~\cite{rumelhart1986learning}. 
Then, we describe the initialization strategies of the Transformer models to avoid the vanishing and exploding gradients problem.

\subsection{Back-Propagation}
Back-propagation passes the gradients of the loss function from the top layer to the bottom layer through the network.
Here, to avoid the vanishing and exploding gradients problem in deep neural networks, the scale of the gradients must be kept constant throughout the model.
Let us consider a layer $\bm{y} =g(\bm{x}) \quad (\bm{y} \in \mathbb{R}^{d_{\mathrm{out}}}, \bm{x} \in \mathbb{R}^{d_{\mathrm{in}}})$.
$\mathcal{L}$ denotes the loss, and $\bm{\delta} \in \mathbb{R}^{d_{\mathrm{out}}}$ denotes the gradient of the loss with respect to the output $\frac{\partial \mathcal{L}}{\partial \bm{y}}$.
To keep the scale of the gradients before and after the layer, a layer $g$ must satisfy the condition,
%can be written as
\begin{align}
E\left[\left\|\frac{\partial \mathcal{L}}{\partial \bm{x}}\right\|^2\right] = %\approx 
E\left[\left\|
\frac{\partial \bm{y}}{\partial \bm{x}}\bm{\delta}
\right\|^2\right] =
E\left[\left\|\bm{\delta}\right\|^2\right].
\label{eq:goal}
\end{align}
Back-propagation is a chain of differentiation. Therefore, the scale of the gradients in the entire model is maintained when each layer in the model meets this requirement.

\subsection{Initialization Strategies of Transformer}
\label{ssec:init_strategy}
\paragraph{Embedding Scaling.}
$\sigma_e$ plays an essential role in back-propagation through the Transformer layers~\cite{takase2023spike}.
Here, we use the RMSNorm $\bm{y} =\bm{\gamma}_\mathrm{LN} \odot \frac{\sqrt{d}\bm{x}}{\sqrt{\|\bm{x}\|^2}}$% +\beta_\mathrm{LN}$
~\cite{rmsnorm} as the layer normalization, %\footnote{The LayerNorm adds the centering operation before the RMSNorm, which does not affect the discussion.}
where $\bm{\gamma}_\mathrm{LN}$ is a parameter, $d$ is the number of dimensions, and $\odot$ indicates the Hadamard product. Back-propagation through RMSNorm is
\[
\frac{\partial \bm{y}}{\partial \bm{x}} =\sqrt{\frac{d}{\|\bm{x}\|^2}} \left(\bm{I} - \frac{\bm{x}\bm{x}\top}{\|\bm{x}\|^2}\right)\mathrm{diag}(\bm{\gamma}_\mathrm{LN}),
\]
where $\mathrm{diag}(\cdot)$ is a diagonal matrix and $\bm{I}$ is an identity matrix.
Because $\sqrt{\frac{d}{\|\bm{x}\|^2}}$ is the inverse of the standard deviation of $\bm{x}$ if the mean of $\bm{x}$ is zero, the standard deviation of $\bm{x}$ affects the norm of the gradients.
The standard deviation of the embedding matrix $\sigma_e$ influences the standard deviation of the input in RMSNorm through the residual connections (Equation~\ref{eq:residual_forward}).
Thus, to avoid the vanishing and exploding gradients problem, $\sigma_e$ should be set to 1.

%Instead of directly setting $\sigma_e =1$, which is much larger than other parameters, \citet{takase2023spike} presented two previous studies. 
On the basis of the above discussion, \citet{takase2023spike}  presented two previous studies achieving a standard deviation of 1 for $\bm{x}$ without directly setting $\sigma_e =1$. %, which is much larger than the other $\sigma_{\cdot}$. 
The first way multiplies the output of the embedding layer by a constant $1/\sigma_e$. This technique was introduced in the original Transformer~\cite{transformer} but was deleted from the implementations. The second way adds the layer normalization to the top of the embedding layer~\cite{le-scao-etal-2022-language}. 

\paragraph{Residual Scaling.}
$\sigma_o$ and $\sigma_d$ are also important factors for stable training. % because of the residual connection. 
The residual scaling technique was introduced to Transformer by GPT-2~\cite{gpt-2} without explanation.
Here, we present a theoretical analysis \cite{taki2017deep} originally designed for ResNet~\cite{resnet} while modifying it for
Transformer. %Although GPT-2~\cite{gpt-2} used this technique, they did not presented any explanations.
The analysis in a formal form is presented in Appendix~\ref{append:formal}.

\iffalse
The transformation of the self-attention layer and the feed-forward layer can be written as 
\begin{align}
\bm{y} =f(\mathrm{LN}(\bm{x})) + \bm{x},
\label{eq:residual_forward}
\end{align}
where LN indicates the RMSNorm.
Here, we assume Pre-LN, which is reported to stabilize the training~\cite{liu-etal-2020-understanding}.
\fi

The back-propagation through Equation~\ref{eq:residual_forward} is
\begin{align}
\frac{\partial \mathcal{L}}{\partial \bm{x}} = \frac{\partial\mathcal{L}}{\partial \bm{y}} \frac{\partial \bm{y}}{\partial \bm{x}} =\bm{\delta} \left(\frac{\partial f(\mathrm{LN}(\bm{x)})}{\partial \bm{x}} +\bm{I}\right).
\label{eq:residual}
\end{align}
Let $s^2$ be $E\left[\left\|\frac{\partial f(\mathrm{LN}(\bm{x)})_i}{\partial \bm{x}}\right\|^2\right]$.
%Equation~\ref{eq:residual} means that 
Thus, a residual connection causes an $(s^2+1)$-fold increase in the squared norm of the gradient $E\left[\left\|\frac{\partial \mathcal{L}}{\partial \bm{x}}\right\|^2\right]$. As a result, the gradient explodes exponentially with respect to the depth of layers throughout the propagation.
%式\ref{eq:goal}で目標として示したように，勾配爆発を抑える観点からは，$s$が大きな値となることは好ましくない．
This exponential increase is unacceptable for LLMs consisting of many Transformer blocks.

%The popular libraries providing large language model pre-training, Megatron-DeepSpeed\footnote{\url{https://github.com/microsoft/Megatron-DeepSpeed}}, GPT-NeoX~\cite{gpt-neox-library}, and llm-foundry\footnote{\url{https://github.com/mosaicml/llm-foundry}}, 
To alleviate this problem, the residual scaling multiplies $\sigma_o$ and $\sigma_d$ by $\frac{1}{\sqrt{2N}}$ since the model has $2N$ residual connections. This multiplication achieves $E[s^2] =\mathcal{O}\left(\frac{1}{2N}\right)$, and the scale of the exploding gradient $(s^2+1)^{2N}$ converges to Napier's constant $e$ in the limit $N\rightarrow \infty$. This avoids an exponential explosion with respect to $N$. %This initialization method is noted by GPT-2~\cite{gpt-2} without referring the discussion above.
%, implemented the initialization method that $\sigma_o$ and $\sigma_d$ are multiplied by $\frac{1}{\sqrt{2N}}$ compared to other parameters. %ここで，$N$はthe Transformerの層数である．この背景には以下の理由がある．Self-Attention変換（$\bm{W}_k, \bm{W}_q$によるアテンション行列の導出，アテンション行列と$\bm{W}_vx$の積，その$\bm{W}_o$による線形変換）を$f$を書く．このとき，Self-Attention層では

\section{Existing Methods and Their Problems}
\label{sec:problem}
Here, we review two of the existing initialization methods and their problems. The methods are summarized in Table~\ref{tab:init_list}.

\subsection{He Initialization}
He initialization~\cite{He_2015_ICCV} is one of the most popular initialization methods for neural networks.
It is designed to keep the scale of the gradients constant throughout the network to meet the requirement of Equation~\ref{eq:goal}. %, to keep the scale of the gradient constant through the linear layer, 
In the case of a linear layer $\bm{y} =\bm{Wx} \quad (\bm{y} \in \mathbb{R}^{d_{\mathrm{out}}}, \bm{x} \in \mathbb{R}^{d_{\mathrm{in}}}, \bm{W} \in \mathbb{R}^{d_{\mathrm{out}} \times d_{\mathrm{in}}})$, the requirements can be written as 
\[
E\left[\left\|\frac{\partial \mathcal{L}}{\partial \bm{x}}\right\|^2\right] =
E\left[\left\|\bm{W}^\top \bm{\delta}\right\|^2\right] =
\mathrm{Var}\left[\left\|\bm{W}^\top \bm{\delta}\right\|\right]
\]
\[
= d_\mathrm{in}\mathrm{Var}\left[\bm{W}\right]E\left[\left\|\bm{\delta}\right\|^2\right]
=E\left[\left\|\bm{\delta}\right\|^2\right].
\]
Thus, the parameter $\bm{W} \in \mathbb{R}^{d_{\mathrm{out}} \times d_{\mathrm{in}}}$ must be initialized with the standard deviation $\sigma =\frac{1}{\sqrt{d_\mathrm{in}}}$.
Note that the numerator, called the gain, is determined depending on the activation function. We assume the identity function in the above discussion for simplicity. For ReLU activation, the gain is $\sqrt{2}$.

\subsection{Small Initialization}
Small initialization~\cite{nguyen-salazar-2019-transformers} is based on empirical findings that a small standard deviation leads to stable training. 
It sets a common small standard deviation $\sqrt{\frac{2}{5d}}$ for all parameters except for the $1/\sqrt{2N}$ scaling of $\sigma_o$ and $\sigma_d$. Here, we should note that $\sqrt{\frac{2}{5d}}$ is the standard deviation which Xavier initialization~\cite{pmlr-v9-glorot10a} specifies for $\bm{W}_u$ and $\bm{W}_d$, and it is the smallest standard deviation among all of the parameters in the Transformer layers.

%わざわざ書かなくてもいいかな
%Small Initializationによって学習が安定する理由に，以下の2点が挙げられる．
%(1) 全てのパラメータを共通の学習率で更新する場合，スケールの大きいパラメータでは学習が進みづらく，スケールの小さいパラメータでは学習が壊れやすい．そのため，できるだけ全てのパラメータの標準偏差を揃えることは重要である．
%(2) He Initializationよりも小さいパラメータを採用することで，Residual Block内部では勾配が減衰し，パラメータの更新量が小さくなる．モデルの全体ではResidual Connectionがあることから勾配が減衰しないため，Residual Block内部での勾配の減衰はモデル全体での勾配減衰を起こさない．

\subsection{Problems}
\label{ssec:problems}
%本研究では，Small Initializationの課題として，以下の2点を指摘する．(1) Residual結合に起因して$\sigma_o, \sigma_d$の$\sqrt{1/2N}$倍が必要であり，全てのパラメータのスケールは統一できてはいない．(2) 勾配をあえて減衰させる戦略は，学習を抑制する懸念がある．

%This paper indicates non-uniformity of the norm of the parameters as one of the causes of loss spikes.
\iffalse
To avoid loss spikes, we can set the learning rate $\mu$ to a small value according to $\bm{W}_d$. However, its value is too small for $\bm{W}_u$, which results in a slow loss decrease.
On the contrary, if $\mu$ is set according to $\bm{W}_u$, it is large for $\bm{W}_d$ and induces loss spikes.
\fi

\textcolor{black}{
Although the He and Small initializations with the embedding and residual scaling stabilize the training, they often cause loss spikes, as shown at the top of Figure~\ref{fig:13B_init_loss}.
\iffalse
\textcolor{black}{
Since the parameter $\bm{W}$ is updated according to 
$\bm{\delta} \bm{x}^\top$ (Equation \ref{eq:backprop_w}),
the update $\Delta \bm{W}$ is determined by $\bm{\delta}$ and $\bm{x}$. %, and $\bm{W}$ has only a slight influence via $\bm{\delta}$. 
Here, $\bm{x}$ is not affected by $\bm{W}$ by definition, and $\bm{\delta}$ is calculated by the back-propagation from the top layer (\textit{i.e.,} a chain of Equation \ref{eq:backprop_x} from the differentiation of $\mathcal{L}$ at the prediction layer). Thus, $\bm{W}$ has little influence on $\bm{\delta} \bm{x}^\top$ only via the value of $\mathcal{L}$.
Because $\Delta \bm{W}$ is defined regardless of the norm of $\bm{W}$, in the parameter $\bm{W}$ with the smaller norm, the norm of $\Delta \bm{W}$ is large compared to that of the parameter itself. Thus, we argue that the non-uniformity of the parameter norms leads unstable training.}
\fi
%Because the Transformer model is non-linear function, the small $\bm{W}$ does not mean the small gradient $\frac{\partial \mathcal{L}} {\partial \bm{W}}$. Moreover, 
Deep neural networks are designed to keep the scale of the gradients constant throughout the model. % to avoid the vanishing and exploding gradients problem. %Therefore, in the case of parameters whose norm is relatively smaller than that of the others, the update $\|\Delta \bm{W}\|/\| \bm{W}\|$ is relatively larger than the others. 
Therefore, in the parameters whose norm is smaller than that of the others, the update ratios $\|\Delta \bm{W}\|/\| \bm{W}\|$ are larger.
Because the update ratio indicates the magnitude of the effect of the update on the parameter, parameters with large update ratios are fragile.
%we argue that the non-uniformity of the parameter norms leads unstable training.
}

The bottom of Figure~\ref{fig:13B_init_loss} shows the update ratios in the last feed-forward layer: $\|\Delta \bm{W}_d\|/\| \bm{W}_d\|$ and $\|\Delta \bm{W}_u\|/\| \bm{W}_u\|$. 
The update ratio of $\bm{W}_d$ is larger than that of $\bm{W}_u$ because the residual scaling multiplies $\|\bm{W}_d\|$ by $1/\sqrt{2N}$ (in the 13B model, $1/\sqrt{2N} \approx 0.11$). The update ratio of $\bm{W}_d$ is especially large at the very beginning. After the pre-training on 1B tokens with some loss spikes, it stays within a certain range. However, it is still much larger than that of $\bm{W}_u$. After the largest loss spike occurs, the update ratio of $\bm{W}_d$ gets closer to that of $\bm{W}_u$. 
Therefore, we consider that uneven and large update ratios can cause loss spikes, and we can mitigate loss spikes by regulating them.
%In other words, the training becomes more stable as the update ratio in $\bm{W}_d$ becomes more stable.

\stepcounter{footnote}
\begin{table}[t!]
\centering
    %\scalebox{0.75}{
    \small
    \tabcolsep3pt
		\begin{tabular}{l|cccccc} \toprule
                & \multicolumn{2}{c}{He} & \multicolumn{2}{c}{Small} & \multicolumn{2}{c}{WeSaR} \\
                & Gate & Weight & Gate & Weight & Gate & Weight \\ \midrule
                $\bm{W}_e$ & 1 & $\sqrt{\frac{1}{d}}$ & 1& $\sqrt{\frac{2}{5d}}$ & 1 & $\sigma$ \\ 
                $\bm{W}_k$ & N/A & $\sqrt{\frac{1}{d}}$ & N/A & $\sqrt{\frac{2}{5d}}$ & $\sqrt{\frac{1}{d}}$ & $\sigma$\\ 
                $\bm{W}_q$ & N/A & $\sqrt{\frac{1}{d}}$ & N/A & $\sqrt{\frac{2}{5d}}$ & $\sqrt{\frac{1}{d}}$ & $\sigma$\\ 
                $\bm{W}_v$ & N/A & $\sqrt{\frac{1}{d}}$ & N/A & $\sqrt{\frac{2}{5d}}$ & $\sqrt{\frac{1}{d}}$ & $\sigma$\\ 
                $\bm{W}_o$ & N/A & $\sqrt{\frac{1}{2Nd}}$ & N/A & $\sqrt{\frac{2}{10Nd}}$ & $\sqrt{\frac{1}{2Nd}}$ & $\sigma$\\ 
                $\bm{W}_u$ & N/A & $\sqrt{\frac{1}{d}}$ & N/A & $\sqrt{\frac{2}{5d}}$ & $\sqrt{\frac{1}{d}}$ & $\sigma$\\ 
                $\bm{W}_d$ & N/A & $\sqrt{\frac{2}{8Nd}}$ & N/A & $\sqrt{\frac{2}{10Nd}}$ & $\sqrt{\frac{2}{8Nd}}$ & $\sigma$\\ 
                $\bm{W}_p$ & N/A & $\sqrt{\frac{1}{d}}$ & N/A& $\sqrt{\frac{2}{5d}}$ & $\sqrt{\frac{1}{d}}$ & $\sigma$ \\ \bottomrule
        \end{tabular}%}
	\caption{Standard deviations of initialization methods before and after the gate\protect\footnotemark[\thefootnote]. We assume that He and Small initializations use embedding scaling~\cite{transformer,takase2023spike}. The proposed method initializes all parameters with a common $\sigma$. We adopt the popular setting where $d_{\mathrm{out}}$ of $\bm{W}_u$ and $d_{\mathrm{in}}$ of $\bm{W}_o$ are 4$d$ and $d_{\mathrm{in}}$ and $d_{\mathrm{out}}$ of the other parameters are $d$.}
 \label{tab:init_list}
\end{table}

\footnotetext[\thefootnote]{We approximate the gain of the activation function used in the feed-forward layer to that of ReLU (\textit{i.e.,} $\sqrt{2}$).}

\section{Proposed Method}
\label{sec:proposed}
We propose WeSaR as a way to %address the above problems. 
%WeSaR is designed to 
meet the two conflicting aforementioned requirements: (i) the criteria of any initialization method designed to avoid the vanishing and exploding gradients problem, as discussed in \S\ref{ssec:init_strategy}, and (ii) the common scales of all parameters to keep stable and uniform update ratios for mitigating loss spikes, as discussed in \S\ref{ssec:problems}. 
In addition to stabilizing the training, WeSaR enables a hyperparameter setting that achieves a rapid decrease in loss.

\subsection{Initialization via Reparameterization}
\label{ssec:proposd}
We consider a situation where the parameter $\bm{W}_{\cdot}$ is initialized according to $\mathcal{N} (0, \sigma_{\cdot}^2)$.
Here, the proposed method initializes $\bm{W}_{\cdot}$ by using a common standard deviation $\sigma$ among all parameters and uses $\bar{\bm{W}}_{\cdot}$ instead of the original $\bm{W}_{\cdot}$ inside the model,
\[
\bm{W}_{\cdot} \sim \mathcal{N} (0, \sigma^2)
\]
\[
\bar{\bm{W}}_{\cdot} =\frac{\sigma_{\cdot}}{\sigma} \bm{W}_{\cdot} = \alpha_{\cdot} \bm{W}_{\cdot},
\]
%That is, instead of $\bm{W}_{\cdot}$, we register the common-scale $\bar{\bm{W}}_{\cdot}$ to the model.
where $\sigma$ is a hyperparameter, and $\alpha_{\cdot}, \bm{W}_{\cdot}$ are trainable parameters. %For simplification and computational efficiency, we train $\alpha_{\cdot} \in \mathbb{R}$ instead of $\sigma_{\cdot}$ and it is initialized to $\frac{\sigma_{\cdot}}{\sigma}$. 
The gate parameter $\alpha_{\cdot}$ %reparameterizes $\bar{\bm{W}}_{\cdot}$ and 
is initialized to $\frac{\sigma_{\cdot}}{\sigma}$. We call $\bm{W}_{\cdot}$ an actual parameter and $\bar{\bm{W}}_{\cdot} =\alpha_{\cdot} \bm{W}_{\cdot}$ a virtual parameter.
%We train $\bar{\bm{W}_{\cdot}}$ instead of $\bm{W}_{\cdot}$ and train the scale parameter of each matrix $\bm{W}_{\cdot}$, $\alpha_{\cdot}$, instead of $\sigma_{\cdot}$ in the implementation.}
%Then, the parameter $\bar{\bm{W}}_{\cdot}$ is used in the model after the multiplication by $\frac{\sigma_{\cdot}}{\sigma}$.

%By introducing the gate parameters to all parameter matrices, WeSaR meets two conflicting requirements: (i) the criteria of any initialization method designed to avoid the vanishing and exploding gradients problem and (ii) the arbitrary common scales of all parameters to keep stable and uniform update ratio for mitigating loss spikes. WeSaR assigns the former requirement to the virtual parameters and the later to the actual parameters.
Beyond introducing the gate parameters to all parameter matrices, WaSAR is designed to initialize the actual parameters with uniform standard deviations $\sigma$ while aligning the standard deviations of the virtual parameter $\sigma_{\cdot}$ to the criteria of the initialization methods by adjusting the gate parameter $\alpha_{\cdot}$.
%This reparameterization technique enables initialization of the actual parameters with uniform standard deviations %without affecting the behavior of the model.
%because we can align the standard deviations of the virtual parameter to the criteria of the initialization methods by adjusting the gate parameter $\alpha$.
Therefore, WeSaR eliminates the non-uniformity of $\|\bm{W}_{\cdot}\|$ and $\|\Delta \bm{W}_{\cdot} \|/\|\bm{W}_{\cdot}\|$.
The effect of WeSaR is shown at the bottom of Figure~\ref{fig:13B_init_loss}. Because $\bm{W}_d$ and $\bm{W}_u$ are initialized equally, their update ratios are comparable and stable during training.

%In the implementation, $\frac{\sigma_{\cdot}}{\sigma}$ is registered as the trainable parameter $\alpha_{\cdot} \in \mathbb{R}$ and it is initialized to $\frac{\sigma_{\cdot}}{\sigma}$. 
Because just one trainable parameter is added to each parameter matrix $\bm{W}_{\cdot} \in \mathrm{R}^{d_\mathrm{out} \times d_\mathrm{in}}$, WeSaR has little effect on the number of trainable parameters and the training cost. Moreover, it has no effect on the inference because the gate parameter can be merged after the training.

We can align the backbone initialization of WeSaR to any existing initialization methods.
In this paper, we adopt He initialization $\frac{\mathrm{gain}}{\sqrt{d_{\mathrm{in}}}}$ with the embedding and residual scaling for the virtual parameter $\alpha_{\cdot} \bm{W}_{\cdot}$ to avoid gradient decay throughout the Transformer layers.

%Beyond the simple extension of embedding scaling~\cite{transformer, takase2023spike}, 
%By introducing the gate parameters to all parameter matrices, WeSaR meets two conflicting requirements: (i) the criteria of any initialization method designed to avoid the vanishing and exploding gradients problem and (ii) the arbitrary common scales of all parameters to keep stable and uniform update ratio for mitigating loss spikes. WeSaR assigns the former requirement to the virtual parameters and the later to the actual parameters.

%Because we can set $\sigma_{\cdot}$ to any values, we can align the WeSaR to any existing initialization methods without affecting actual parameter $\bar{\bm{W}}_{\cdot}$. 
%$\sigma$ is a hyperparameter.

%We also found that the trainable $\alpha$ reduces the update of the actual parameters during training. We consider that it makes the training stable. We will discuss this topic in the experimental section.

\subsection{Theoretical Justification}
Here, we explain that WeSaR does not affect the training dynamics of Transformer.
We assume that the optimizer is Adam~\cite{adam} because of its benefits to Transformer~\cite{zhang2020adaptive, pan2022toward, zhang2024transformers}.
Let us consider a parameter update $\Delta \bm{W}_t$ at step $t$.
The update of Adam is
\begin{align}
    \label{eq:adam}
    \Delta \bm{W}_t =\mu_t \frac{\bm{M}_t}{\sqrt{\bm{V}_t}},
\end{align}
where $\bm{M}_t$ is the exponential moving average of the gradient $\frac{\partial \mathcal{L}}{\partial \bm{W}}$, $\bm{V}_t$ is that of the squared gradient, and $\mu_t$ is the learning rate.

Because of 
\[
\frac{\partial \mathcal{L}}{\partial \bm{W}_{\cdot}} =\frac{\partial \mathcal{L}}{\partial \bar{\bm{W}}_{\cdot}} \frac{\partial \bar{\bm{W}}_{\cdot}}{\partial \bm{W}_{\cdot}} = \frac{\sigma_{\cdot}}{\sigma} \frac{\partial \mathcal{L}}{\partial \bar{\bm{W}}_{\cdot}}, 
\]
the gradient is multiplied by $\frac{\sigma_{\cdot}}{\sigma}$ through the gate. 
From the definition of Adam (Equation \ref{eq:adam}),  the Adam states $\bm{M}_t$ and $\sqrt{\bm{V}_t}$ are multiplied by $\frac{\sigma_{\cdot}}{\sigma}$ equally, and thus the reparameterization does not affect the parameter update $\mu_t \frac{\bm{M}_t}{\sqrt{\bm{V}_t}}$.
Therefore, the parameter update is independent of $\sigma_{\cdot}$ if we use Adam. 

%仮に$x$や$\bm{\delta}$が定数倍されたとき，$\frac{\partial \mathcal{L}}{\partial \bm{W}} =\bm{\delta} x^\top$より，勾配も定数倍される．が，更新量については$\bm{M}_t$と$\sqrt{\bm{V}_t}$によって定数倍が相殺されるため，影響を受けない．このように，時刻に非依存な勾配のスケールはAdamでは相殺され，全てのパラメータが$\mathcal{O}(\mu_t)$で学習される．

That is, WeSaR relieves the actual parameters and their update of the restriction with respect to $\sigma_{\cdot}$ that is specified in order to avoid the vanishing and exploding gradients problem. Secondarily, different from the existing methods that define the standard deviations as functions of $d$, we can determine the standard deviation of the actual parameters independently of $d$, because the gate $\alpha$ undertakes the dependence on $d$.
%the Transformerモデルにおける勾配爆発・消失を抑制するために規定されるモデルパラメータのスケール$\sigma_{\cdot}$が，パラメータ$\bar{\bm{W}}_{\cdot}$にもパラメータの更新量$\Delta \bar{\bm{W}}_{\cdot}$にも現れない．これは，LLMの学習から$\sigma_{\cdot}$に関する制約を実質的に取り除いたと解釈できる．
%言い換えれば，ゲートパラメータを導入することで，ゲートパラメータにモデル全体の勾配を安定化させるための役割を担わせ，パラメータの初期値に学習が高速に進める役割を担わせることができる．
%さらに副次的に，初期値のスケールを$d$に依存させる既存研究と異なり，$d$への依存もゲート$\alpha$が担うことで，モデルサイズに非依存に初期値のスケールを定めることができる．%実際に，初期値の適切なスケールはモデルサイズに非依存であることを数値実験によって確認する．

%これは既存研究に回せる
%文献~\cite{NEURIPS2022_ae0cba71}は層を経てトークン表現が近寄っていき隠れ状態行列のランクが落ちていくrank collapseを指摘し，$\bm{W}_o$と$\bm{W}_d$による線形変換の直後に$1/\sqrt{N}$をかけることを提案した．本研究は，埋め込み行列の定数スケーリング（small initなら$\sqrt{5d/2}$）を含めた全てのパラメータにスケーリングを導入し，任意のスケールでパラメータを初期化することを可能にした点が新しい．

\iffalse
Furthermore, the advantage of the proposed methods the decoupling of the scale of the propagation to the lower layers and the propagation to the parameter:
\[
\frac{\partial \mathcal{L}}{\partial \bm{x}} =\bm{W}^\top \bm{\delta} =\alpha \bar{\bm{W}} \bm{\delta}
\]
\[
\frac{\partial \mathcal{L}}{\partial \bm{W}} =\bm{\delta} x^\top.
\]
\fi

\subsection{Hyperparameter Setting}
\label{ssec:propose_init}
Here, we explain the hyperparameter setting that enables a stable and rapid loss decrease. %, which WeSaR achieves.
%In addition to the common standard deviation $\sigma$ that introduced by WeSaR, we can set the learning rage and the batch size to accelerate training due to the stability.
Different from conventional initialization methods, WeSaR can set the common standard deviation $\sigma$ to an arbitrary value.
In addition, the stability afforded by WeSaR enables us to set the learning rate and batch size to accelerate training.

\paragraph{Standard Deviation $\sigma$.}
%As discussed in \S\ref{ssec:proposd}, we can set the standard deviation of the actual parameters to an arbitrary value. 
In this paper, we set to $\sigma^2 =~$\textrm{4e-5}, unless otherwise mentioned. This setup corresponds to $d =10,000$ in the Small initialization criteria $\sqrt{\frac{2}{5d}}$. That is, our $\sigma$ setup is smaller than those of conventional setups\footnote{Even in LLaMA3 70B, $d =8192$~\cite{llama3modelcard}.}. We can expect a rapid decrease in loss with the same learning rate because of the large parameter update $\Delta \bm{W}$ relative to the parameter $\bm{W}$ itself. \citet{zhang-etal-2019-improving} confirmed the preference to a smaller standard deviation in the Transformer models, which justifies our setup.

\paragraph{Learning rate.}
Because WeSaR enables stable training, we can increase the learning rate from the conventional values (an order of 1e-4). Here, we set it to 1e-3.

\paragraph{Batch size.}
In the conventional pre-training of an LLM, the batch size is set to a large value (e.g., 4M tokens) to avoid loss spikes. 
We can decrease the batch size for a rapid loss decrease if the training is stable. However, the batch size has to be large enough in order to pre-train the model efficiently on large numbers of GPUs, as is commonly done when pre-training LLMs. 
%small batch size is incompatible with multiprocessing based on a large number of GPUs. 
Thus, we set the batch size to 1M tokens.

\begin{table}[t!]
\centering
    %\scalebox{0.75}{
    \small
    \tabcolsep3pt
		\begin{tabular}{l|ccc} \toprule
                & 130M & 1.3B & 13B \\
                \midrule
                \# Param. & 134.1M & 1,339.1M & 12,911.0M \\
                Hidden Size $d$ & 768 & 2048 & 5120 \\   
                \# Layer $N$ & 12 & 24 & 40 \\   
                \# Attention Head & 12 & 16 & 40 \\ 
                \bottomrule
        \end{tabular}%}
	\caption{Model configuration.}
 \label{tab:model_config}
\end{table}

\begin{table}[t!]
\centering
    %\scalebox{0.75}{
    \small
    \tabcolsep3pt
		\begin{tabular}{l|cc} \toprule
                & Rapid Setting & Stable Setting \\ \midrule
                Batch Size [tokens] & 1M & 4M \\
                Learning rate $\mu$ & 1e-3 & 5e-4 \\
                Warmup Steps & 100 & 2000 \\
                Gradient Clipping Threshold & \multicolumn{2}{c}{1} \\
                Weight decay & \multicolumn{2}{c}{0.01} \\  
                Z-loss & \multicolumn{2}{c}{1e-4}  \\
                \bottomrule
        \end{tabular}%}
	\caption{Training configuration.}
 \label{tab:optim_config}
\end{table}

\section{Experimental Evaluation}
\label{sec:experiments}

\subsection{Experimental Setup}
\label{ssec:model_config}
We pre-trained the 130M, 1.3B, and 13B models on the basis of the configuration listed in Table~\ref{tab:model_config}.
The model architecture was based on LLaMA~\cite{touvron2023llama}, except for the feed-forward layer with gelu activation.
Our experiments mainly focused on the 1.3B models.
The training was based on the hyperparameters listed in Table~\ref{tab:optim_config}.
There were two settings for the learning rate, batch size, and warmup steps: One was a conventional setting emphasizing on a stable training; the other emphasized a rapid decrease in loss.
We used perplexity as a metric.
Appendix~\ref{append:experimental_setup} describes the detailed configuration.

\subsection{Dataset}
We sampled 30B tokens from RefinedWeb~\cite{penedo2023refinedweb} and used them as the pre-training corpus. %Because of our computational resources, we sampled 30B tokens from it.
\citet{hoffmann2022training} found that the optimal pre-training corpus size is roughly 20 tokens per model parameter. Thus, 30B tokens were sufficient for our main experiments using 1.3B models.
For the 13B models, we investigated the behavior in the first 1/10th of the training.
For the evaluation, we used LAMBADA~\cite{paperno-etal-2016-lambada} and WikiText~\cite{merity2017pointer}.

\subsection{Compared Models}
As a baseline, we trained the model with the most popular method, \textit{i.e.,} \textbf{Small initialization}.

In addition, we compared the proposed method with the three reparameterization methods listed in Table~\ref{tab:reparam_list}. Because all methods have their own motivation, we discuss the detailed difference in Appendix~\ref{append:reparam_compare}. In short, the difference from the former two methods is efficiency because WeSaR does not conduct any normalization. From the last method, WeSaR reparameterizes all parameters and sets a common small value to the standard deviations of all parameters.

\begin{table}[t!]
\centering
    %\scalebox{0.75}{
    \small
    \tabcolsep3pt
		\begin{tabular}{l|cccc} \toprule
            Method & Weights & Train & Norm & Scale \\ \midrule
                Weight Normalization & all & $\checkmark$ & $\checkmark$ & by-row \\ 
                $\sigma$Reparam & all & $\checkmark$ & $\checkmark$ & by-matrix \\
                Residual Scaling & $\bm{W}_o$, $\bm{W}_d$ & & & by-matrix \\  \midrule
                WeSaR & all & $\checkmark$ & & by-matrix \\                 
                \bottomrule
        \end{tabular}%}
	\caption{Comparison of reparameterization methods.
``Weights'' means the reparameterized weight matrices. "Train" means that each method uses trainable gate parameters. ``Norm'' means that each method uses reparameterization via weight-based normalization. ``Scale'' means the unit of scaling in the reparameterization.}
 \label{tab:reparam_list}
\end{table}

\paragraph{Weight Normalization.}
Weight Normalization~\cite{NIPS2016_ed265bc9} was proposed to decouple the length of the weight vectors from their direction. It conducts L2 normalization and scaling of each row of the parameter matrix $\bm{w} \in \mathbb{R}^{d_\mathrm{in}}$ as $
\bar{\bm{w}} = \frac{\alpha}{\|\bm{w}\|}\bm{w}$.
%We also modified it to efficient matrix-wise reparameterization $\bar{\bm{W}} =\frac{\alpha}{\|\bm{W}\|} \bm{W}$. 

\paragraph{$\mathrm{\sigma}$Reparam.}
$\mathrm{\sigma}$Reparam~\cite{pmlr-v202-zhai23a} was proposed to control the spectral norm (\textit{i.e.,} the maximum singular value) of the parameter for stable Transformer training. It conducts spectral normalization~\cite{miyato2018spectral} and scaling of the parameter matrix $\bm{W} \in \mathrm{R}^{d_\mathrm{out} \times d_\mathrm{in}}$:
$
\bar{\bm{W}} = \frac{\alpha}{\|\bm{W}\|_2}\bm{W},
$
where $\|\bm{W}\|_2$ is the spectral norm.
The original $\sigma$Reparam adopts Post-LN; and we tried both Post-LN and the more popular Pre-LN.

\paragraph{Residual Scaling as Reparameterization.}
\citet{NEURIPS2022_ae0cba71} overcomes the limitation of the $(1/\sqrt{2N})$-fold multiplications of $\sigma_o$ and $\sigma_d$ caused by the residual connection (Equation \ref{eq:residual_forward}). It modifies the residual connection to 
$
\bm{y} =\frac{1}{\sqrt{2N}}f(\mathrm{LN}(\bm{x)}) + \bm{x}.
$
Different from the original residual scaling, which changes the standard deviations, this equation can be viewed as a reparameterization of $\bm{W}_o$ and $\bm{W}_d$ because of its linearity.

\paragraph{Setup.} For Weight Normalization and $\sigma$Reparam, which reparameterize all parameters, we tuned $\sigma^2$ in $\{1, 4, 16, 64, 256 \}$e-5 and set the initial $\alpha$ to the values defined by each method. Because residual scaling does not reparameterize all of the parameters and does not specify a backbone initialization method, we chose the He and Small initializations. All methods used embedding scaling because  \citet{takase2023spike} confirmed its benefit.

\subsection{Results and Discussion}

\begin{table}[t!]
\centering
    %\scalebox{0.75}{
    \small
    \tabcolsep3pt
		\begin{tabular}{l|l|cc} \toprule
                & &  WikiText & LAMBADA \\
                \midrule
                \multirow{3}{*}{\rotatebox[origin=c]{90}{130M}} 
                & Small Init. (Rapid) & 26.57 & 33.56 \\
                & Small Init. (Stable) & 37.68 & 40.41 \\ \cmidrule{2-4}
                & WeSaR & \textbf{25.07} & \textbf{31.89} \\ \midrule
                \multirow{3}{*}{\rotatebox[origin=c]{90}{1.3B}} & Small Init. (Rapid) & 16.55 & 26.29 \\ 
                & Small Init. (Stable) & 21.44 & 28.81 \\ \cmidrule{2-4}
                & WeSaR & \textbf{14.51} & \textbf{22.87} \\
                \midrule
                \multirow{3}{*}{\rotatebox[origin=c]{90}{13B}} & Small Init. (Rapid) & 12.72 & 21.79 \\
                & Small Init. (Stable) & 18.66 & 25.34 \\ \cmidrule{2-4} 
                & WeSaR & \textbf{12.05} & \textbf{21.57} \\
                \bottomrule
        \end{tabular}%}
	\caption{Main results.}
 \label{tab:main_result}
\end{table}

\begin{figure}[t!]
\centering
		\includegraphics[width=0.8\linewidth]{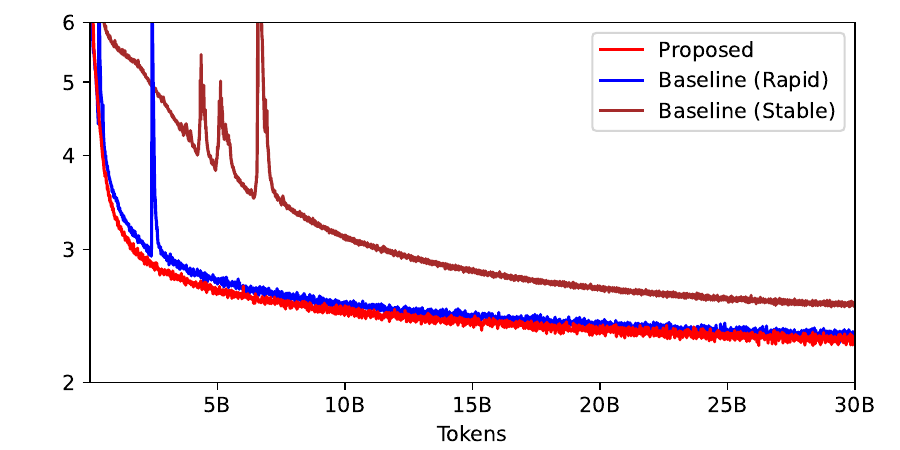}
		\caption{Loss of 13B models during training.}
        \label{fig:13B_loss}
\end{figure}

\paragraph{Main results.}
Table~\ref{tab:main_result} shows the main results. WeSaR outperformed the widely used Small initialization. Figure~\ref{fig:13B_init_loss} and \ref{fig:13B_loss} show the decrease in loss of the 13B models at the beginning of and over the whole training, respectively. We found that WeSaR achieved stable training, whereas Small initialization caused loss spikes. Moreover, under the hyperparameter setting that aimed to stabilize training, Small initialization still caused loss spikes and eventually had higher (\textit{i.e.,} worse) perplexity due to the small learning rate and large batch size. As well, due to the lower learning rate, the stable setting took more steps until reaching stable states without loss spikes.
Thus, we used the rapid hyperparameter setting in the following experiments. 
The loss decreases for the 130M and 1.3B models are shown in Appendix~\ref{append:loss_decrease}.

\begin{table*}[t!]
\centering
    %\scalebox{0.75}{
    \small
		\begin{tabular}{l|ccccc} \toprule
                & WikiText & LAMBADA & Time & \# Param. & Best $\sigma^2$ \\
                \midrule
                Small Init. & 20.64 (0.52) & 29.50 (0.53) & 18.88~~~~~~~~~~~~~~~~ & 1,339.1M & N/A \\ \midrule
                Weight Normalization & 18.87 (0.59) & \underline{27.69} (0.86) & 21.27~(+12.6\%) & 1,339.6M & 16e-5 \\ \midrule
                %WeightNorm. w./ matrix reparam. & 18.26 (0.31) & \underline{27.67} (0.28)  & 21.34~(+13.0\%) & 1,339.1M & 4e-5 \\ \midrule
                $\mathrm{\sigma}$Reparam w./ Pre-LN & 25.26 (1.65) & 30.74 (0.74) & 20.06~(+6.25\%) & 1,339.1M & 64e-5 \\ 
                $\mathrm{\sigma}$Reparam w./ Post-LN & 23.64 (1.03) & 30.56 (0.89) & 20.09~(+6.39\%) & 1,339.1M & 16e-5 \\ \midrule
                Residual Scaling w./ He & 23.15 (0.37) & 31.03 (0.20) & 19.19~(+1.66\%) & 1,339.1M & N/A \\ 
                Residual Scaling w./ Small & 23.56 (1.03) & 30.78 (0.35) & 19.18~(+1.58\%) & 1,339.1M & N/A \\ \midrule
                WeSaR & \textbf{17.74} (0.05) & \textbf{27.52} (0.28) & 19.25~(+1.95\%) & 1,339.1M & 4e-5 \\
                \bottomrule
        \end{tabular}%}
	\caption{Comparison of reparameterization methods in five runs based on 10B tokens. Mean and standard deviation are listed. The best method is in bold, and the methods within one standard deviation are underlined.}
 \label{tab:reparam_result_std}
\end{table*}

\begin{figure}[t!]
\centering
		\includegraphics[width=0.8\linewidth]{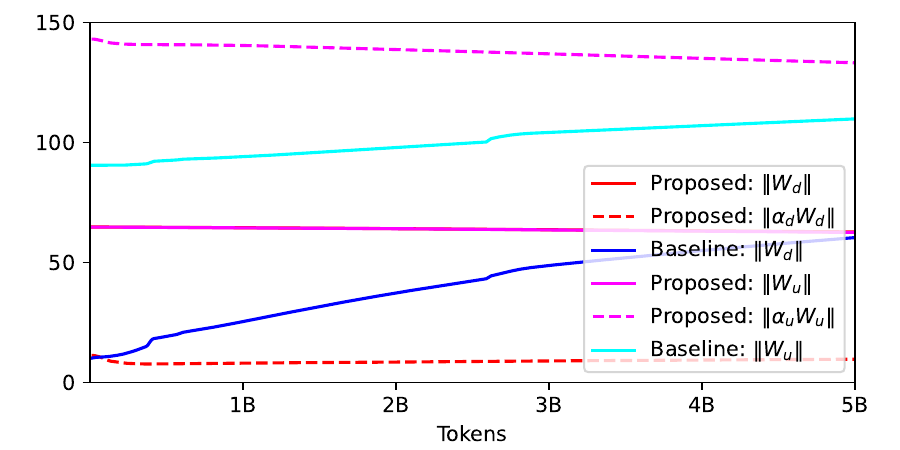}
		\caption{Norm of parameters $\| \bm{W}_d\|$ and $\| \bm{W}_u\|$ in the last layer at the beginning of the training. $\| \bm{W}_d\|$ and $\| \bm{W}_u\|$ of the proposed method overlap.}
        \label{fig:13B_init_param}
\end{figure}

\begin{table}[t!]
\centering
    %\scalebox{0.75}{
    \small
    \tabcolsep3pt
		\begin{tabular}{l|cc} \toprule
                & WikiText & LAMBADA \\
                \midrule
                Small Init. & 16.55 & 26.29 \\
                He Init. & 16.70 & 26.50 \\ \midrule
                WeSaR (w./ He Init.) &  \textbf{14.51} & \textbf{22.87} \\
                \quad w./ Small Init. & 15.91 & 24.37 \\ 
                \quad w./ fixed $\alpha$ & 15.21 & 25.61 \\ 
                %\quad Small $\rightarrow$ He & & & \\
                \bottomrule
        \end{tabular}%}
	\caption{Ablation studies.} \label{tab:ablation_result}
\end{table}

\begin{table*}[t!]
\centering
    %\scalebox{0.75}{
    \footnotesize
        \tabcolsep2pt
		\begin{tabular}{l|cccccccccc|ccc} \toprule
Dataset & BoolQ & \multicolumn{2}{c}{CB} & COPA & MultiRC & \multicolumn{2}{c}{ReCoRD }& RTE & WiC & WSC & \multicolumn{3}{c}{Total} \\ \midrule
Metric & ACC & ACC & F1 & ACC & ACC & EM & F1 & ACC & ACC & ACC & ACC & EM & F1 \\ \midrule
Small Init. & \underline{60.28} & 32.14 & 22.26 & \underline{73.00} & 46.12 & 73.23 & 73.93 & \underline{53.43} &\textbf{ 50.00} & 40.38 & 51.73 & 73.23 & 73.65 \\ \midrule
Weight Normalization & 58.27 & \underline{42.86} & 25.13 & \underline{69.00} & \textbf{57.32} & 75.11 & 75.83 & \textbf{57.76} & \textbf{50.00} & 36.54 & \underline{57.18} & 75.11 & 75.55 \\ \midrule
$\mathrm{\sigma}$Reparam w./ Pre-LN  & \underline{61.19} & \textbf{48.21} & 28.78 & 66.00 & 50.08 & 68.32 & 69.02 & \underline{52.71} & \textbf{50.00} & \underline{44.23} & 54.16 & 68.32 & 68.79 \\ 
$\mathrm{\sigma}$Reparam w./ Post-LN  & 57.65 & \underline{46.43} & 26.63 & \underline{68.00} & 52.83 & 71.69 & 72.41 & \underline{53.79} & \textbf{50.00} & \textbf{52.88} & 54.46 & 71.69 & 72.16 \\ \midrule
Residual Scaling w./ He  & 57.80 & 33.93 & 33.28 & \underline{69.00} & \underline{57.10} & 72.17 & 72.82 & \underline{54.15} & \textbf{50.00} & \underline{51.92} & 56.70 & 72.17 & 72.60 \\
Residual Scaling w./ Small & \underline{60.73} & 33.93 & 23.04 & 66.00 & \underline{57.08} & 71.32 & 72.01 & 51.62 & \textbf{50.00} & 42.31 & \underline{57.51} & 71.32 & 71.74  \\ \midrule
WeSaR & \textbf{61.62} & \underline{41.07} & \textbf{38.54} & \textbf{76.00} & \underline{56.81} & \textbf{76.68} & \textbf{77.37} & 50.54 & \underline{48.75} & \underline{44.23} & \textbf{57.73} & \textbf{76.68} & \textbf{77.16} \\
%std & 0.85 & 6.63 & & 4.29 & 0.71 & 0.42 & 0.42 & 3.01 & 1.98 & 4.89 & 0.51 & 0.42 & \\
%Best-std & & 6.73 &   & &0.71& & &  2.97 & & 4.92 & & & \\
        \bottomrule
        \end{tabular}%}
	\caption{Evaluation of 1.3B models on downstream tasks. The best method is in bold, and the methods within one standard deviation are underlined.}
    \label{tab:downstream}
\end{table*}

\textcolor{black}{
\paragraph{Why does the reparameterization stabilize training?}
The bottom of Figure~\ref{fig:13B_init_loss} shows that, during the training using Small initialization, $\|\Delta \bm{W}_d \|/ \|\bm{W}_d\|$ was large at the very beginning of the training and became small and stable after the loss spikes occurred. However, the proposed method kept $\|\Delta \bm{W}_d \|/ \|\bm{W}_d\|$ and $\|\Delta \bm{W}_u \|/ \|\bm{W}_u\|$ in a certain range during the training, which led to stable training.
The update ratios in other parameters are shown in Appendix~\ref{append:relative_update}.
}

\textcolor{black}{
To investigate the reason for this remarkable difference, we analyzed the values of $\|\bm{W}_d\|$ and $\|\bm{W}_u\|$ in the last layer during training. As shown in Figure~\ref{fig:13B_init_param},
$\|\bm{W}_d\|$ and $\|\bm{W}_u\|$ of Small initialization became larger during training because of the small initial values.
To achieve such large change in $\bm{W}_d$ and $\bm{W}_u$, the parameter update should be also large enough. %\Delta \bm{W}_d$ and $\Delta \bm{W}_u$ are required to have large values. 
Therefore, the update ratios of Small initialization were larger and more unstable than those of WeSaR. A large update is especially harmful to $\bm{W}_d$ due to the non-uniformity, which causes the training to become unstable.
}

\textcolor{black}{
Although the virtual parameters $\alpha_d \bm{W}_d$ and $\alpha_u \bm{W}_u$ of WeSaR changed their norms during training, WeSaR assigned the role of changing the norm to the gate parameter $\alpha_d$ and $\alpha_u$. Therefore, the norm of the actual parameters $\|\bm{W}_d\|$ and $\|\bm{W}_u\|$ did not change by much. This nearly constant scale of the actual parameters contributed to the stability.
}

\paragraph{Is the reparameterization effective?}
Table~\ref{tab:ablation_result} shows the results of the ablation studies.
Among the existing methods, Small initialization outperformed He initialization. He initialization also caused loss spikes. % (See Appendix~\ref{append:loss_decrease}). 
Thus, as \citet{nguyen-salazar-2019-transformers} confirmed, Small initialization is more suitable than He initialization for pre-training LLMs.

However, %because WeSaR %decouples the effect of the back-propagation to lower layers and to actual parameters by reparameterization, 
%relieves the parameters of the demand by the back-propagation, 
%implemented the initialization method in the virtual parameter $\alpha \bar{W}$,
He initialization outperformed Small initialization as a backbone initialization method of WeSaR. We consider that He initialization is suitable for propagating the gradients to lower layers, although a small standard deviation (\textit{e.g.,} Small initialization) is suitable as the parameter itself. %. In addition, as we confirmed in the experiments with the existing methods, a small standard deviation (\textit{e.g.,} Small Initialization) is suitable for the actual parameter $\bar{W}$. 
The advantage of WeSaR is that it sets the standard deviations of the actual parameter to smaller values, while it sets the norm of the virtual parameter to a sufficient value for the back-propagation. %This result justifies the decoupling strategy enabled by the proposed method.

Also, in relation to discussed with Figure~\ref{fig:13B_init_param}, the trainability of the gate parameter $\alpha$ contributes to the model performance.

\paragraph{Does WeSaR outperform the existing reparameterization methods?}
We compared WeSaR with the existing reparameterization methods, shown in Table~\ref{tab:reparam_result_std}. 
In pilot experiments, we confirmed that the pre-training on 10B tokens is sufficient to rank the methods. Thus, we conducted five runs of each method with 10B tokens and report the means and the standard deviations. The single runs on the full 30B tokens are described in Appendix~\ref{append:full_run}. 

WeSaR achieved a lower (\textit{i.e.,} better) perplexity on average and smaller (\textit{i.e.,} more stable) standard deviations than Weight Normalization. 
In addition, Weight Normalization took the longest time.
This is because that it calculates the back-propagation through the normalization, different from the other methods. 
%In addition, the original Weight Normalization uses the vector-wise reparameterization that requires more time consumption and parameters than our matrix-wise reparameterization.
%achieved comparable perplexity with WeSaR. However, its time consumption is largest among all methods. Regarding the original vector-wise reparameterization, the more efficient matrix-wise reparameterization, which requires fewer parameters like WeSaR, outperformed the original vector-wise reparameterization. 
%Weight Normalization achieved comparable perplexity to WeSaR. 
We confirmed that our simple reparameterization without normalization is efficient and effective for LLM's pre-training.

\iffalse
The proposed method outperformed Weight Normalization in LAMBADA. 
However, the vector-wise Weight Normalization outperformed all methods in WikiText.
Here, compared to the matrix-wise Weight Normalization, the vector-wise one reduced the perplexity in WikiText in which the models achieved smaller perplexity, but increased it in LAMBADA in which it did larger perplexity.
Therefore, we consider that the vector-wise reparameterization tends to cause over-fitting to the trained domain although there is little difference in the number of parameters.
We concluded that our matrix-wise reparameterization is both efficient and effective for LLM initialization. Also, from the comparison of the proposed method and the matrix-wise Weight Normalization, we consider that scaling parameters without normalization is sufficient for the LLM pre-training.
\fi

Moreover, WeSaR outperformed $\sigma$Reparam. Whereas $\sigma$Reparam controls the attention entropy for stability, WeSaR stabilizes the training by sharing the standard deviations of all of the parameters even without spectral normalization. In addition, we consider that setting the initial standard deviation to the criteria of He initialization achieved a more rapid decrease in loss than did setting the initial maximum singular value to 1. %We also note that a difference of 0.46 hour can occur as an error because the Pre-LN and Post-LN do not differ in total amount of computation.

Third, WeSaR outperformed residual scaling in terms of perplexity. Because residual scaling only reparameterizes $\bm{W}_o$ and $\bm{W}_d$, we consider that the relief of all of the parameters from the requirements by the back-propagation, which also results in smaller standard deviations than in the conventional setting, is important for a stable and rapid decrease in loss.

\paragraph{Is WeSaR effective on downstream tasks?}
To confirm the effectiveness of WeSaR on downstream tasks, we evaluated the compared models on the SuperGLEU dataset~\cite{wang2019superglue} via lm-evaluation-harness~\cite{eval-harness}. 
We used BoolQ~\cite{clark2019boolq}, CB~\cite{demarneffe:cb}, COPA~\cite{roemmele2011choice}, MultiRC~\cite{khashabi2018looking}, ReCoRD~\cite{zhang2018record}, RTE~\cite{dagan2006pascal, bar2006second,giampiccolo2007third, bentivogli2009fifth}, WiC~\cite{pilehvar2018wic}, and WSC~\cite{levesque2011winograd} with the official metrics in lm-evaluation-harness.
We did not conduct fine-tuning and report the results with 3-shot in-context learning.

Table~\ref{tab:downstream} lists the results. In addition to the perplexity as language modeling, the model pre-trained with WeSaR outperformed the compared models on the downstream tasks on average.

\begin{table}[t!]
\centering
    %\scalebox{0.75}{
    \small
        \tabcolsep3pt
		\begin{tabular}{l|l|cc} \toprule
                & & WikiText & LAMBADA \\
                \midrule
                \multirow{6}{*}{\rotatebox[origin=c]{90}{130M ($d=768$)}} & $\sigma^2=16\textrm{e-5}~(d=5000) $ & 28.64 & 36.52 \\
                & $\sigma^2=4\textrm{e-5}~(d=10000) $ & 25.07 & \textbf{31.89} \\
                & $\sigma^2=1\textrm{e-5}~(d=20000) $ & \textbf{24.51} & 33.46 \\ \cmidrule{2-4}
                & $\mu=2\textrm{e-3}$ & 24.55 & 33.15 \\
                & $\mu=1\textrm{e-3}$ & 25.07 & \textbf{31.89} \\ 
                & $\mu=5\textrm{e-4}$ & \textbf{24.50} & 33.25 \\ \midrule
                \multirow{6}{*}{\rotatebox[origin=c]{90}{1.3B ($d=2048$)}} & $\sigma^2=16\textrm{e-5}~(d=5000) $ & 16.37 & 26.05 \\
                & $\sigma^2=4\textrm{e-5}~(d=10000) $ & \textbf{14.51} & \textbf{22.87} \\
                & $\sigma^2=1\textrm{e-5}~(d=20000) $ & 14.85 & 24.19 \\ \cmidrule{2-4}
                & $\mu=2\textrm{e-3}$ & 14.67 & 24.02 \\
                & $\mu=1\textrm{e-3}$ & \textbf{14.51} & \textbf{22.87} \\ 
                & $\mu=5\textrm{e-4}$ & 15.98 & 25.59 \\ \midrule
                \bottomrule
        \end{tabular}%}
	\caption{Robustness versus standard deviation $\sigma$ and learning rate $\mu$. The parentheses in the second columns indicate the number of dimensions measured against the criteria of Small initialization.}
    \label{tab:robustness}
 \end{table}

\paragraph{Are the hyperparameter settings robust to changes in model size?}
Table~\ref{tab:robustness} clarifies the robustness with respect to the model size.

Here, we observed that the $\sigma^2 =4\textrm{e-5}$ setting outperformed the other settings 
in the 1.3B model experiments, while the $\sigma^2 =4\textrm{e-5}$ and 1e-5 settings achieved comparable performance in the 130M model experiments. 
Although there remains room for tuning the hyperparameters, we found that the optimal standard deviations are not necessarily proportional to the dimension size $d$, different from the conventional setup; a larger model does not always prefer a smaller standard deviation. This is because the back-propagation to lower layers must depend on $d$ and the proposed method assigns the role of ensuring this dependence to the gate parameter. 
Second, regarding the learning rate, we confirmed that WeSaR achieves stable training even with a higher rate (order of 1e-3) than that of conventional settings (order of 1e-4).

\iffalse
\paragraph{Down-streaming taskにおける性能は高いか？Fine-Tuningでも有効か？}
1.3Bモデルだけでもやっておきたい．

\paragraph{モデルは広さと深さどちらが重要か？}
Optional. 極端なd,Lの設定でもWeSaRが頑健に動くことを示せるとよさそう．合わせて，どれが性能がいいかを考察．

\paragraph{他の初期化と組み合わせられるか？}
Opiional.
\fi

\section{Related Work}
\paragraph{Loss spikes.}
PaLM~\cite{palm} and OPT~\cite{opt} found the loss spike phenomenon and used a simple strategy against it that restarts the training from an earlier checkpoint and skips batches that may have caused the spike.
GLM~\cite{glm} found that the abnormal gradients of the embedding layer usually cause spikes and proposed to shrink the gradients of $\bm{W}_e$. 
\citet{NEURIPS2022_aac02401} and \citet{pmlr-v202-zhai23a} argued that large context lengths and abnormal attention behavior lead to spikes.
\citet{molybog2023theory} indicated that the Adam optimizer, which assumes time-domain independence of gradients, induces loss spikes.
\citet{takase2023spike} presented embedding scaling~\cite{transformer} and LayerNorm on the top of the embedding layer~\cite{le-scao-etal-2022-language} by focusing the differentiation of the layer normalization.
The causes of loss spikes are still under intense discussion. We clarified that one of the causes is the non-uniformity of the parameter norms and provided a method to address this issue.

\paragraph{Residual scaling.} 
The $(1/\sqrt{2N})$-fold initialization of $\sigma_d, \sigma_o$ was first proposed in LLM studies by GPT-2~\cite{gpt-2}. %, they did not provide any mathematical explanations. 
Apart from Transformer, \citet{taki2017deep,NEURIPS2018_d81f9c1b,zhang2018residual} %, NEURIPS2019_e520f70a, pmlr-v97-allen-zhu19a} 
presented a weight scaling for ResNet~\cite{resnet} together with a mathematical justification. 
%One of the difference of the residual connections in the Transformer from those in ResNet is the multiple transformation in a residual branch. Thus, 
%More recently, \citet{wang2022deepnet, pmlr-v119-huang20f, NEURIPS2022_ae0cba71} proposed $(2N)^{-1/4}$-fold initialization of $\sigma_d, \sigma_u, \sigma_v, \sigma_o$. It seems to lead stable training because the non-uniformity is relieved than $(1/\sqrt{2N})$-fold initialization. 
Some recent studies have proposed weight scaling for Transformer and have given theoretical analyses, including $\mathcal{O}(N^{-1/4})$-fold scaling of $\bm{W}_v, \bm{W}_o$ \cite{pmlr-v119-huang20f},
$\mathcal{O}(N^{-1/2})$ of $\bm{W}_o, \bm{W}_d$ 
 as reparameterization \cite{NEURIPS2022_ae0cba71}, and $\mathcal{O}(N^{-1/4})$ of $\bm{W}_v, \bm{W}_o, \bm{W}_u$, and $\bm{W}_d$
\cite{wang2022deepnet}.
%It seems to lead stable training because the non-uniformity is relieved than $(1/\sqrt{2N})$-fold initialization. 
%However, the popular libraries providing large language model pre-training, Megatron-DeepSpeed\footnote{\url{https://github.com/microsoft/Megatron-DeepSpeed}}, GPT-NeoX~\cite{gpt-neox-library}, and llm-foundry\footnote{\url{https://github.com/mosaicml/llm-foundry}}, implemented GPT-2's initialization method. 
We have extended this line of work to the novel reparameterization method. Although we used the most popular GPT-2's strategy for the initial scale, we can use any of the scaling strategies described above.
%%文献~\cite{wang2022deepnet, pmlr-v119-huang20f, NEURIPS2022_ae0cba71}はで$W_v$と$W_o$は$(2N)^{-\frac{1}{4}}$倍ずつ行うことを提案している．
%文献~\cite{wang2022deepnet, pmlr-v119-huang20f}は，ResNetの一連の研究とは独立に，the TransformerにおけるResidual結合に起因する問題を指摘し，初期化手法を提案した．彼らは，一回の更新での損失関数値の変化を$\Theta(\mu)$とすることを目標にし，$W_v$と$W_o$を$\mathcal{O}(N^{-\frac{1}{4})}$で初期化することを提案した．
%%\citet{wang2022deepnet}は一回の更新での損失関数値の変化を$\Theta(\mu)$とするため，$8N^{-\frac{1}{4}}$としResidual結合時に$2N^{-\frac{1}{4}}$倍の操作を行うことを提案した．
%しかし，これらの論文の公開にもかかわらず，GPT-NeoXやllm-foundryといった主要な事前学習ライブラリでは$W_o, W_d$の$1/\sqrt{2N}$倍の実装が採用されている．自然言語処理の分野で$1/\sqrt{2N}$の必要性の理解がその根拠とともに進んでいるとは言えない．

%比較対象が増えかねないし，必要なものは引用しているし，わざわざ書かなくていいかな
\iffalse
\paragraph{Reparameterization methods.} 
Reparameterization methods are mainly studied for sparsification and quantization of neural network. Former methods conduct decomposition of parameter matrices~\cite{hoge} or introduce hypernetwork that outputs parameter matrices. %LoRA~\cite{lora} is one of the most popular methods in LLM studies. 
Later methods scale the parameters to express them in low-precision matrix~\cite{hoge}.
This paper shows that the reparameterization methods is effective to control an update ratio $\|\Delta W\| / \|W\|$ and to mitigate loss spikes.
\fi

\paragraph{Initialization methods.} Some studies have determined the initial scale of the parameters with a prior optimization phase before the pre-training~\cite{NEURIPS2019_876e8108,NEURIPS2021_88ae6372,NEURIPS2022_7886b9ba,Bingham_Miikkulainen_2023}. Our method can use them as the backbone initialization instead of He initialization.

\section{Conclusion}
Loss spikes are a fundamental issue in pre-training of LLMs because they increase the pre-training cost and degrade the performance of the model. To address this problem, we identified one of the causes as the non-uniformity of the norm of the model parameters. %, which is required to avoid the vanishing and exploding gradients problem. 
We proposed a novel reparameterization method, WeSaR, that addresses the non-uniformity problem by adjusting the gate parameter to the required scale and initializing the actual parameters with a common standard deviation. 
WeSaR not only stabilizes the pre-training, but also accelerates the pre-training by setting a standard deviation smaller than in the conventional setting. 
Experimental results showed that WeSaR outperformed the compared methods, and the parameters and their update ratios were stable during pre-training. %Due to the stability, we can set the hyperparameters while aiming to a rapid loss decrease. 

The use of LLMs has been spreading. We believe this study to be a significant contribution that increases both the efficiency of the LLM's pre-training and the effectiveness of the pre-trained LLMs.

\section*{Limitations}
The proposed method and the presented theoretical analysis focus on one aspect of the loss spike problem and does not solve it entirely. In the experiments, we used various techniques designed for stable training: warmup, Adam $\beta_2 =0.95$, gradient clipping, weight decay, and Z-loss. We do not insist that such techniques are no longer required. For example, in pilot experiments with the 1.3B model, we found that no warmup or no gradient clipping training achieved higher perplexity due to unstable behavior at the very beginning of the training. We argue that there is no silver bullet against loss spikes and that we should address this issue with a combination of techniques, including WeSaR.

Another limitation is the restriction of the computational resources. For example, our experiments investigated the behavior of the models with up to 13B parameters.
Moreover, we did not use SWiGLU activation~\cite{shazeer2020glu} in the feed-forward layers, as has been done in popular LLMs, \textit{e.g.}, PaLM~\cite{palm} and LLaMA~\cite{touvron2023llama}. However, we note that the effectiveness of SWiGLU remains controversial in the community: \citet{narang-etal-2021-transformer} has a positive opinion, and \citet{allen2024physics} a negative one.
%12*8*(3+4)+60*8*(3+2+5*5+4+7*5/3)+40*64*3 =30272
%$39.33 * 30272 /8 =$148,824
In spite of these restrictions, our experiments showed the effectiveness of WeSaR on a standard Transformer architecture. Our experiments took 30,272 GPU hours on H100 totally. This would cost \$148,824 if the experiments were conducted on Amazon Web Service in June, 2024. 
We believe that our findings based on the intensive experiments shed new light on LLMs.

\bibliography{j_yourrefs}

\begin{thebibliography}{61}
\providecommand{\natexlab}[1]{#1}

\bibitem[{AI@Meta(2024)}]{llama3modelcard}
AI@Meta. 2024.
\newblock \href {https://github.com/meta-llama/llama3/blob/main/MODEL_CARD.md} {Llama 3 model card}.

\bibitem[{Allen-Zhu and Li(2024)}]{allen2024physics}
Zeyuan Allen-Zhu and Yuanzhi Li. 2024.
\newblock Physics of language models: Part 3.3, knowledge capacity scaling laws.
\newblock \emph{arXiv preprint arXiv:2404.05405}.

\bibitem[{Ba et~al.(2016)Ba, Kiros, and Hinton}]{layer_norm}
Jimmy~Lei Ba, Jamie~Ryan Kiros, and Geoffrey~E Hinton. 2016.
\newblock Layer normalization.
\newblock \emph{arXiv preprint arXiv:1607.06450}.

\bibitem[{Bar~Haim et~al.(2006)Bar~Haim, Dagan, Dolan, Ferro, Giampiccolo, Magnini, and Szpektor}]{bar2006second}
Roy Bar~Haim, Ido Dagan, Bill Dolan, Lisa Ferro, Danilo Giampiccolo, Bernardo Magnini, and Idan Szpektor. 2006.
\newblock The second {PASCAL} recognising textual entailment challenge.

\bibitem[{Bentivogli et~al.(2009)Bentivogli, Dagan, Dang, Giampiccolo, and Magnini}]{bentivogli2009fifth}
Luisa Bentivogli, Ido Dagan, Hoa~Trang Dang, Danilo Giampiccolo, and Bernardo Magnini. 2009.
\newblock The fifth {PASCAL} recognizing textual entailment challenge.

\bibitem[{Bingham and Miikkulainen(2023)}]{Bingham_Miikkulainen_2023}
Garrett Bingham and Risto Miikkulainen. 2023.
\newblock \href {https://doi.org/10.1609/aaai.v37i6.25836} {Autoinit: Analytic signal-preserving weight initialization for neural networks}.
\newblock \emph{Proceedings of the AAAI Conference on Artificial Intelligence}, 37(6):6823--6833.

\bibitem[{Brown et~al.(2020)Brown, Mann, Ryder, Subbiah, Kaplan, Dhariwal, Neelakantan, Shyam, Sastry, Askell et~al.}]{gpt-3}
Tom~B Brown, Benjamin Mann, Nick Ryder, Melanie Subbiah, Jared Kaplan, Prafulla Dhariwal, Arvind Neelakantan, Pranav Shyam, Girish Sastry, Amanda Askell, et~al. 2020.
\newblock Language models are few-shot learners.
\newblock \emph{arXiv preprint arXiv:2005.14165}.

\bibitem[{Chowdhery et~al.(2023)Chowdhery, Narang, Devlin, Bosma, Mishra, Roberts, Barham, Chung, Sutton, Gehrmann et~al.}]{palm}
Aakanksha Chowdhery, Sharan Narang, Jacob Devlin, Maarten Bosma, Gaurav Mishra, Adam Roberts, Paul Barham, Hyung~Won Chung, Charles Sutton, Sebastian Gehrmann, et~al. 2023.
\newblock Palm: Scaling language modeling with pathways.
\newblock \emph{Journal of Machine Learning Research}, 24(240):1--113.

\bibitem[{Clark et~al.(2019)Clark, Lee, Chang, Kwiatkowski, Collins, and Toutanova}]{clark2019boolq}
Christopher Clark, Kenton Lee, Ming-Wei Chang, Tom Kwiatkowski, Michael Collins, and Kristina Toutanova. 2019.
\newblock {B}ool{Q}: Exploring the surprising difficulty of natural yes/no questions.
\newblock In \emph{Proceedings of NAACL-HLT 2019}.

\bibitem[{Dagan et~al.(2006)Dagan, Glickman, and Magnini}]{dagan2006pascal}
Ido Dagan, Oren Glickman, and Bernardo Magnini. 2006.
\newblock The {PASCAL} recognising textual entailment challenge.
\newblock In \emph{Machine learning challenges. evaluating predictive uncertainty, visual object classification, and recognising tectual entailment}, pages 177--190. Springer.

\bibitem[{Dauphin and Schoenholz(2019)}]{NEURIPS2019_876e8108}
Yann~N Dauphin and Samuel Schoenholz. 2019.
\newblock \href {https://proceedings.neurips.cc/paper_files/paper/2019/file/876e8108f87eb61877c6263228b67256-Paper.pdf} {Metainit: Initializing learning by learning to initialize}.
\newblock In \emph{Advances in Neural Information Processing Systems}, volume~32.

\bibitem[{De~Marneffe et~al.(2019)De~Marneffe, Simons, and Tonhauser}]{demarneffe:cb}
Marie-Catherine De~Marneffe, Mandy Simons, and Judith Tonhauser. 2019.
\newblock {The CommitmentBank}: Investigating projection in naturally occurring discourse.
\newblock To appear in proceedings of Sinn und Bedeutung 23. Data can be found at https://github.com/mcdm/CommitmentBank/.

\bibitem[{Gao et~al.(2024)Gao, Tow, Abbasi, Biderman, Black, DiPofi, Foster, Golding, Hsu, Le~Noac'h, Li, McDonell, Muennighoff, Ociepa, Phang, Reynolds, Schoelkopf, Skowron, Sutawika, Tang, Thite, Wang, Wang, and Zou}]{eval-harness}
Leo Gao, Jonathan Tow, Baber Abbasi, Stella Biderman, Sid Black, Anthony DiPofi, Charles Foster, Laurence Golding, Jeffrey Hsu, Alain Le~Noac'h, Haonan Li, Kyle McDonell, Niklas Muennighoff, Chris Ociepa, Jason Phang, Laria Reynolds, Hailey Schoelkopf, Aviya Skowron, Lintang Sutawika, Eric Tang, Anish Thite, Ben Wang, Kevin Wang, and Andy Zou. 2024.
\newblock \href {https://doi.org/10.5281/zenodo.12608602} {A framework for few-shot language model evaluation}.

\bibitem[{Giampiccolo et~al.(2007)Giampiccolo, Magnini, Dagan, and Dolan}]{giampiccolo2007third}
Danilo Giampiccolo, Bernardo Magnini, Ido Dagan, and Bill Dolan. 2007.
\newblock The third {PASCAL} recognizing textual entailment challenge.
\newblock In \emph{Proceedings of the ACL-PASCAL workshop on textual entailment and paraphrasing}, pages 1--9. Association for Computational Linguistics.

\bibitem[{Glorot and Bengio(2010)}]{pmlr-v9-glorot10a}
Xavier Glorot and Yoshua Bengio. 2010.
\newblock \href {https://proceedings.mlr.press/v9/glorot10a.html} {Understanding the difficulty of training deep feedforward neural networks}.
\newblock In \emph{Proceedings of the Thirteenth International Conference on Artificial Intelligence and Statistics}, volume~9 of \emph{Proceedings of Machine Learning Research}, pages 249--256.

\bibitem[{Hanin and Rolnick(2018)}]{NEURIPS2018_d81f9c1b}
Boris Hanin and David Rolnick. 2018.
\newblock \href {https://proceedings.neurips.cc/paper_files/paper/2018/file/d81f9c1be2e08964bf9f24b15f0e4900-Paper.pdf} {How to start training: The effect of initialization and architecture}.
\newblock In \emph{Advances in Neural Information Processing Systems}, volume~31.

\bibitem[{He et~al.(2015)He, Zhang, Ren, and Sun}]{He_2015_ICCV}
Kaiming He, Xiangyu Zhang, Shaoqing Ren, and Jian Sun. 2015.
\newblock Delving deep into rectifiers: Surpassing human-level performance on imagenet classification.
\newblock In \emph{Proceedings of the IEEE International Conference on Computer Vision}.

\bibitem[{He et~al.(2016)He, Zhang, Ren, and Sun}]{resnet}
Kaiming He, Xiangyu Zhang, Shaoqing Ren, and Jian Sun. 2016.
\newblock Deep residual learning for image recognition.
\newblock In \emph{Proceedings of the IEEE conference on computer vision and pattern recognition}, pages 770--778.

\bibitem[{Hoffmann et~al.(2022)Hoffmann, Borgeaud, Mensch, Buchatskaya, Cai, Rutherford, Casas, Hendricks, Welbl, Clark et~al.}]{hoffmann2022training}
Jordan Hoffmann, Sebastian Borgeaud, Arthur Mensch, Elena Buchatskaya, Trevor Cai, Eliza Rutherford, Diego de~Las Casas, Lisa~Anne Hendricks, Johannes Welbl, Aidan Clark, et~al. 2022.
\newblock Training compute-optimal large language models.
\newblock \emph{arXiv preprint arXiv:2203.15556}.

\bibitem[{Huang et~al.(2020)Huang, Perez, Ba, and Volkovs}]{pmlr-v119-huang20f}
Xiao~Shi Huang, Felipe Perez, Jimmy Ba, and Maksims Volkovs. 2020.
\newblock \href {https://proceedings.mlr.press/v119/huang20f.html} {Improving transformer optimization through better initialization}.
\newblock In \emph{Proceedings of the 37th International Conference on Machine Learning}, volume 119 of \emph{Proceedings of Machine Learning Research}, pages 4475--4483.

\bibitem[{Kaplan et~al.(2020)Kaplan, McCandlish, Henighan, Brown, Chess, Child, Gray, Radford, Wu, and Amodei}]{kaplan2020scaling}
Jared Kaplan, Sam McCandlish, Tom Henighan, Tom~B Brown, Benjamin Chess, Rewon Child, Scott Gray, Alec Radford, Jeffrey Wu, and Dario Amodei. 2020.
\newblock Scaling laws for neural language models.
\newblock \emph{arXiv preprint arXiv:2001.08361}.

\bibitem[{Khashabi et~al.(2018)Khashabi, Chaturvedi, Roth, Upadhyay, and Roth}]{khashabi2018looking}
Daniel Khashabi, Snigdha Chaturvedi, Michael Roth, Shyam Upadhyay, and Dan Roth. 2018.
\newblock Looking beyond the surface: A challenge set for reading comprehension over multiple sentences.
\newblock In \emph{Proceedings of the 2018 Conference of the North American Chapter of the Association for Computational Linguistics: Human Language Technologies, Volume 1 (Long Papers)}, pages 252--262.

\bibitem[{Kingma and Ba(2015)}]{adam}
Diederik~P. Kingma and Jimmy Ba. 2015.
\newblock \href {http://arxiv.org/abs/1412.6980} {Adam: A method for stochastic optimization}.
\newblock In \emph{ICLR (Poster)}.

\bibitem[{Le~Scao et~al.(2022)Le~Scao, Wang, Hesslow, Bekman, Bari, Biderman, Elsahar, Muennighoff, Phang, Press, Raffel, Sanh, Shen, Sutawika, Tae, Yong, Launay, and Beltagy}]{le-scao-etal-2022-language}
Teven Le~Scao, Thomas Wang, Daniel Hesslow, Stas Bekman, M~Saiful Bari, Stella Biderman, Hady Elsahar, Niklas Muennighoff, Jason Phang, Ofir Press, Colin Raffel, Victor Sanh, Sheng Shen, Lintang Sutawika, Jaesung Tae, Zheng~Xin Yong, Julien Launay, and Iz~Beltagy. 2022.
\newblock \href {https://doi.org/10.18653/v1/2022.findings-emnlp.54} {What language model to train if you have one million {GPU} hours?}
\newblock In \emph{Findings of the Association for Computational Linguistics: EMNLP 2022}, pages 765--782.

\bibitem[{Levesque et~al.(2011)Levesque, Davis, and Morgenstern}]{levesque2011winograd}
Hector~J Levesque, Ernest Davis, and Leora Morgenstern. 2011.
\newblock The {W}inograd schema challenge.
\newblock In \emph{{AAAI} Spring Symposium: Logical Formalizations of Commonsense Reasoning}, volume~46, page~47.

\bibitem[{Li et~al.(2022)Li, Zhang, and He}]{NEURIPS2022_aac02401}
Conglong Li, Minjia Zhang, and Yuxiong He. 2022.
\newblock \href {https://proceedings.neurips.cc/paper_files/paper/2022/file/aac02401755a65904cf977a33136af4a-Paper-Conference.pdf} {The stability-efficiency dilemma: Investigating sequence length warmup for training gpt models}.
\newblock In \emph{Advances in Neural Information Processing Systems}, volume~35, pages 26736--26750.

\bibitem[{Liu et~al.(2020)Liu, Liu, Gao, Chen, and Han}]{liu-etal-2020-understanding}
Liyuan Liu, Xiaodong Liu, Jianfeng Gao, Weizhu Chen, and Jiawei Han. 2020.
\newblock \href {https://doi.org/10.18653/v1/2020.emnlp-main.463} {Understanding the difficulty of training transformers}.
\newblock In \emph{Proceedings of the 2020 Conference on Empirical Methods in Natural Language Processing}, pages 5747--5763.

\bibitem[{Merity et~al.(2017)Merity, Xiong, Bradbury, and Socher}]{merity2017pointer}
Stephen Merity, Caiming Xiong, James Bradbury, and Richard Socher. 2017.
\newblock \href {https://openreview.net/forum?id=Byj72udxe} {Pointer sentinel mixture models}.
\newblock In \emph{International Conference on Learning Representations}.

\bibitem[{Miyato et~al.(2018)Miyato, Kataoka, Koyama, and Yoshida}]{miyato2018spectral}
Takeru Miyato, Toshiki Kataoka, Masanori Koyama, and Yuichi Yoshida. 2018.
\newblock \href {https://openreview.net/forum?id=B1QRgziT-} {Spectral normalization for generative adversarial networks}.
\newblock In \emph{International Conference on Learning Representations}.

\bibitem[{Molybog et~al.(2023)Molybog, Albert, Chen, DeVito, Esiobu, Goyal, Koura, Narang, Poulton, Silva et~al.}]{molybog2023theory}
Igor Molybog, Peter Albert, Moya Chen, Zachary DeVito, David Esiobu, Naman Goyal, Punit~Singh Koura, Sharan Narang, Andrew Poulton, Ruan Silva, et~al. 2023.
\newblock A theory on adam instability in large-scale machine learning.
\newblock \emph{arXiv preprint arXiv:2304.09871}.

\bibitem[{Narang et~al.(2021)Narang, Chung, Tay, Fedus, Fevry, Matena, Malkan, Fiedel, Shazeer, Lan, Zhou, Li, Ding, Marcus, Roberts, and Raffel}]{narang-etal-2021-transformer}
Sharan Narang, Hyung~Won Chung, Yi~Tay, Liam Fedus, Thibault Fevry, Michael Matena, Karishma Malkan, Noah Fiedel, Noam Shazeer, Zhenzhong Lan, Yanqi Zhou, Wei Li, Nan Ding, Jake Marcus, Adam Roberts, and Colin Raffel. 2021.
\newblock \href {https://doi.org/10.18653/v1/2021.emnlp-main.465} {Do transformer modifications transfer across implementations and applications?}
\newblock In \emph{Proceedings of the 2021 Conference on Empirical Methods in Natural Language Processing}, pages 5758--5773.

\bibitem[{Nguyen and Salazar(2019)}]{nguyen-salazar-2019-transformers}
Toan~Q. Nguyen and Julian Salazar. 2019.
\newblock \href {https://aclanthology.org/2019.iwslt-1.17} {Transformers without tears: Improving the normalization of self-attention}.
\newblock In \emph{Proceedings of the 16th International Conference on Spoken Language Translation}, Hong Kong. Association for Computational Linguistics.

\bibitem[{Noci et~al.(2022)Noci, Anagnostidis, Biggio, Orvieto, Singh, and Lucchi}]{NEURIPS2022_ae0cba71}
Lorenzo Noci, Sotiris Anagnostidis, Luca Biggio, Antonio Orvieto, Sidak~Pal Singh, and Aurelien Lucchi. 2022.
\newblock \href {https://proceedings.neurips.cc/paper_files/paper/2022/file/ae0cba715b60c4052359b3d52a2cff7f-Paper-Conference.pdf} {Signal propagation in transformers: Theoretical perspectives and the role of rank collapse}.
\newblock In \emph{Advances in Neural Information Processing Systems}, volume~35, pages 27198--27211.

\bibitem[{Pan and Li(2022)}]{pan2022toward}
Yan Pan and Yuanzhi Li. 2022.
\newblock \href {https://openreview.net/forum?id=Sf1NlV2r6PO} {Toward understanding why adam converges faster than {SGD} for transformers}.
\newblock In \emph{OPT 2022: Optimization for Machine Learning (NeurIPS 2022 Workshop)}.

\bibitem[{Paperno et~al.()Paperno, Kruszewski, Lazaridou, Pham, Bernardi, Pezzelle, Baroni, Boleda, and Fern{\'a}ndez}]{paperno-etal-2016-lambada}
Denis Paperno, Germ{\'a}n Kruszewski, Angeliki Lazaridou, Ngoc~Quan Pham, Raffaella Bernardi, Sandro Pezzelle, Marco Baroni, Gemma Boleda, and Raquel Fern{\'a}ndez.
\newblock \href {https://doi.org/10.18653/v1/P16-1144} {The {LAMBADA} dataset: Word prediction requiring a broad discourse context}.
\newblock In \emph{Proceedings of the 54th Annual Meeting of the Association for Computational Linguistics (Volume 1: Long Papers)}, pages 1525--1534.

\bibitem[{Paszke et~al.(2017)Paszke, Gross, Chintala, Chanan, Yang, DeVito, Lin, Desmaison, Antiga, and Lerer}]{paszke2017automatic}
Adam Paszke, Sam Gross, Soumith Chintala, Gregory Chanan, Edward Yang, Zachary DeVito, Zeming Lin, Alban Desmaison, Luca Antiga, and Adam Lerer. 2017.
\newblock Automatic differentiation in pytorch.

\bibitem[{Penedo et~al.(2023)Penedo, Malartic, Hesslow, Cojocaru, Cappelli, Alobeidli, Pannier, Almazrouei, and Launay}]{penedo2023refinedweb}
Guilherme Penedo, Quentin Malartic, Daniel Hesslow, Ruxandra Cojocaru, Alessandro Cappelli, Hamza Alobeidli, Baptiste Pannier, Ebtesam Almazrouei, and Julien Launay. 2023.
\newblock The refinedweb dataset for falcon llm: outperforming curated corpora with web data, and web data only.
\newblock \emph{arXiv preprint arXiv:2306.01116}.

\bibitem[{Pilehvar and Camacho-Collados(2019)}]{pilehvar2018wic}
Mohammad~Taher Pilehvar and Jose Camacho-Collados. 2019.
\newblock {WiC}: The word-in-context dataset for evaluating context-sensitive meaning representations.
\newblock In \emph{Proceedings of NAACL-HLT}.

\bibitem[{Radford et~al.(2019)Radford, Wu, Child, Luan, Amodei, Sutskever et~al.}]{gpt-2}
Alec Radford, Jeffrey Wu, Rewon Child, David Luan, Dario Amodei, Ilya Sutskever, et~al. 2019.
\newblock Language models are unsupervised multitask learners.
\newblock \emph{{OpenAI} blog}.

\bibitem[{Roemmele et~al.(2011)Roemmele, Bejan, and Gordon}]{roemmele2011choice}
Melissa Roemmele, Cosmin~Adrian Bejan, and Andrew~S. Gordon. 2011.
\newblock Choice of plausible alternatives: An evaluation of commonsense causal reasoning.
\newblock In \emph{2011 AAAI Spring Symposium Series}.

\bibitem[{Rumelhart et~al.(1986)Rumelhart, Hinton, and Williams}]{rumelhart1986learning}
David~E Rumelhart, Geoffrey~E Hinton, and Ronald~J Williams. 1986.
\newblock Learning representations by back-propagating errors.
\newblock \emph{nature}, 323(6088):533--536.

\bibitem[{Salimans and Kingma(2016)}]{NIPS2016_ed265bc9}
Tim Salimans and Durk~P Kingma. 2016.
\newblock \href {https://proceedings.neurips.cc/paper_files/paper/2016/file/ed265bc903a5a097f61d3ec064d96d2e-Paper.pdf} {Weight normalization: A simple reparameterization to accelerate training of deep neural networks}.
\newblock In \emph{Advances in Neural Information Processing Systems}, volume~29.

\bibitem[{Shazeer(2020)}]{shazeer2020glu}
Noam Shazeer. 2020.
\newblock Glu variants improve transformer.
\newblock \emph{arXiv preprint arXiv:2002.05202}.

\bibitem[{Takase et~al.(2023)Takase, Kiyono, Kobayashi, and Suzuki}]{takase2023spike}
Sho Takase, Shun Kiyono, Sosuke Kobayashi, and Jun Suzuki. 2023.
\newblock Spike no more: Stabilizing the pre-training of large language models.
\newblock \emph{arXiv preprint arXiv:2312.16903}.

\bibitem[{Taki(2017)}]{taki2017deep}
Masato Taki. 2017.
\newblock Deep residual networks and weight initialization.
\newblock \emph{arXiv preprint arXiv:1709.02956}.

\bibitem[{Touvron et~al.(2023)Touvron, Lavril, Izacard, Martinet, Lachaux, Lacroix, Rozi{\`e}re, Goyal, Hambro, Azhar et~al.}]{touvron2023llama}
Hugo Touvron, Thibaut Lavril, Gautier Izacard, Xavier Martinet, Marie-Anne Lachaux, Timoth{\'e}e Lacroix, Baptiste Rozi{\`e}re, Naman Goyal, Eric Hambro, Faisal Azhar, et~al. 2023.
\newblock Llama: Open and efficient foundation language models.
\newblock \emph{arXiv preprint arXiv:2302.13971}.

\bibitem[{Vaswani et~al.(2017)Vaswani, Shazeer, Parmar, Uszkoreit, Jones, Gomez, Kaiser, and Polosukhin}]{transformer}
Ashish Vaswani, Noam Shazeer, Niki Parmar, Jakob Uszkoreit, Llion Jones, Aidan~N Gomez, {\L}ukasz Kaiser, and Illia Polosukhin. 2017.
\newblock Attention is all you need.
\newblock In \emph{NIPS}, pages 5998--6008.

\bibitem[{Wang et~al.(2019)Wang, Pruksachatkun, Nangia, Singh, Michael, Hill, Levy, and Bowman}]{wang2019superglue}
Alex Wang, Yada Pruksachatkun, Nikita Nangia, Amanpreet Singh, Julian Michael, Felix Hill, Omer Levy, and Samuel~R. Bowman. 2019.
\newblock Super{GLUE}: A stickier benchmark for general-purpose language understanding systems.
\newblock \emph{arXiv preprint 1905.00537}.

\bibitem[{Wang et~al.(2022)Wang, Ma, Dong, Huang, Zhang, and Wei}]{wang2022deepnet}
Hongyu Wang, Shuming Ma, Li~Dong, Shaohan Huang, Dongdong Zhang, and Furu Wei. 2022.
\newblock Deepnet: Scaling transformers to 1,000 layers.
\newblock \emph{arXiv preprint arXiv:2203.00555}.

\bibitem[{Wolf et~al.(2019)Wolf, Debut, Sanh, Chaumond, Delangue, Moi, Cistac, Rault, Louf, Funtowicz et~al.}]{wolf2019huggingface}
Thomas Wolf, Lysandre Debut, Victor Sanh, Julien Chaumond, Clement Delangue, Anthony Moi, Pierric Cistac, Tim Rault, R{\'e}mi Louf, Morgan Funtowicz, et~al. 2019.
\newblock Huggingface's transformers: State-of-the-art natural language processing.
\newblock \emph{arXiv preprint arXiv:1910.03771}.

\bibitem[{Yang et~al.(2022)Yang, Wang, Yuan, and Lin}]{NEURIPS2022_7886b9ba}
Yibo Yang, Hong Wang, Haobo Yuan, and Zhouchen Lin. 2022.
\newblock \href {https://proceedings.neurips.cc/paper_files/paper/2022/file/7886b9bafe76c52fd568db10ff9772df-Paper-Conference.pdf} {Towards theoretically inspired neural initialization optimization}.
\newblock In \emph{Advances in Neural Information Processing Systems}, volume~35, pages 18983--18995.

\bibitem[{Zeng et~al.(2023)Zeng, Liu, Du, Wang, Lai, Ding, Yang, Xu, Zheng, Xia, Tam, Ma, Xue, Zhai, Chen, Liu, Zhang, Dong, and Tang}]{glm}
Aohan Zeng, Xiao Liu, Zhengxiao Du, Zihan Wang, Hanyu Lai, Ming Ding, Zhuoyi Yang, Yifan Xu, Wendi Zheng, Xiao Xia, Weng~Lam Tam, Zixuan Ma, Yufei Xue, Jidong Zhai, Wenguang Chen, Zhiyuan Liu, Peng Zhang, Yuxiao Dong, and Jie Tang. 2023.
\newblock \href {https://openreview.net/forum?id=-Aw0rrrPUF} {{GLM}-130b: An open bilingual pre-trained model}.
\newblock In \emph{The Eleventh International Conference on Learning Representations}.

\bibitem[{Zhai et~al.(2023)Zhai, Likhomanenko, Littwin, Busbridge, Ramapuram, Zhang, Gu, and Susskind}]{pmlr-v202-zhai23a}
Shuangfei Zhai, Tatiana Likhomanenko, Etai Littwin, Dan Busbridge, Jason Ramapuram, Yizhe Zhang, Jiatao Gu, and Joshua~M. Susskind. 2023.
\newblock \href {https://proceedings.mlr.press/v202/zhai23a.html} {Stabilizing transformer training by preventing attention entropy collapse}.
\newblock In \emph{Proceedings of the 40th International Conference on Machine Learning}, volume 202 of \emph{Proceedings of Machine Learning Research}, pages 40770--40803.

\bibitem[{Zhang and Sennrich(2019)}]{rmsnorm}
Biao Zhang and Rico Sennrich. 2019.
\newblock \href {https://proceedings.neurips.cc/paper_files/paper/2019/file/1e8a19426224ca89e83cef47f1e7f53b-Paper.pdf} {Root mean square layer normalization}.
\newblock In \emph{Advances in Neural Information Processing Systems}, volume~32.

\bibitem[{Zhang et~al.(2019{\natexlab{a}})Zhang, Titov, and Sennrich}]{zhang-etal-2019-improving}
Biao Zhang, Ivan Titov, and Rico Sennrich. 2019{\natexlab{a}}.
\newblock \href {https://doi.org/10.18653/v1/D19-1083} {Improving deep transformer with depth-scaled initialization and merged attention}.
\newblock In \emph{Proceedings of the 2019 Conference on Empirical Methods in Natural Language Processing and the 9th International Joint Conference on Natural Language Processing}, pages 898--909.

\bibitem[{Zhang et~al.(2019{\natexlab{b}})Zhang, Dauphin, and Ma}]{zhang2018residual}
Hongyi Zhang, Yann~N. Dauphin, and Tengyu Ma. 2019{\natexlab{b}}.
\newblock \href {https://openreview.net/forum?id=H1gsz30cKX} {Residual learning without normalization via better initialization}.
\newblock In \emph{International Conference on Learning Representations}.

\bibitem[{Zhang et~al.(2020)Zhang, Karimireddy, Veit, Kim, Reddi, Kumar, and Sra}]{zhang2020adaptive}
Jingzhao Zhang, Sai~Praneeth Karimireddy, Andreas Veit, Seungyeon Kim, Sashank Reddi, Sanjiv Kumar, and Suvrit Sra. 2020.
\newblock Why are adaptive methods good for attention models?
\newblock \emph{Advances in Neural Information Processing Systems}, 33:15383--15393.

\bibitem[{Zhang et~al.(2018)Zhang, Liu, Liu, Gao, Duh, and Durme}]{zhang2018record}
Sheng Zhang, Xiaodong Liu, Jingjing Liu, Jianfeng Gao, Kevin Duh, and Benjamin~Van Durme. 2018.
\newblock {ReCoRD}: Bridging the gap between human and machine commonsense reading comprehension.
\newblock \emph{arXiv preprint 1810.12885}.

\bibitem[{Zhang et~al.(2022)Zhang, Roller, Goyal, Artetxe, Chen, Chen, Dewan, Diab, Li, Lin et~al.}]{opt}
Susan Zhang, Stephen Roller, Naman Goyal, Mikel Artetxe, Moya Chen, Shuohui Chen, Christopher Dewan, Mona Diab, Xian Li, Xi~Victoria Lin, et~al. 2022.
\newblock Opt: Open pre-trained transformer language models.
\newblock \emph{arXiv preprint arXiv:2205.01068}.

\bibitem[{Zhang et~al.(2024)Zhang, Chen, Ding, Li, Sun, and Luo}]{zhang2024transformers}
Yushun Zhang, Congliang Chen, Tian Ding, Ziniu Li, Ruoyu Sun, and Zhi-Quan Luo. 2024.
\newblock Why transformers need adam: A hessian perspective.
\newblock \emph{arXiv preprint arXiv:2402.16788}.

\bibitem[{Zhu et~al.(2021)Zhu, Ni, Xu, Kong, Huang, and Goldstein}]{NEURIPS2021_88ae6372}
Chen Zhu, Renkun Ni, Zheng Xu, Kezhi Kong, W.~Ronny Huang, and Tom Goldstein. 2021.
\newblock \href {https://proceedings.neurips.cc/paper_files/paper/2021/file/88ae6372cfdc5df69a976e893f4d554b-Paper.pdf} {Gradinit: Learning to initialize neural networks for stable and efficient training}.
\newblock In \emph{Advances in Neural Information Processing Systems}, volume~34, pages 16410--16422.

\end{thebibliography}

\appendix

\section{Analysis of Residual Scaling}
\label{append:formal}
Here, we present the detailed explanation of residual scaling.
The back-propagation through Equation~\ref{eq:residual_forward} is
\begin{align}
\frac{\partial \mathcal{L}}{\partial \bm{x}} = \frac{\partial\mathcal{L}}{\partial \bm{y}} \frac{\partial \bm{y}}{\partial \bm{x}} =\bm{\delta} \left(\frac{\partial f(\mathrm{LN}(\bm{x)})}{\partial \bm{x}} +\bm{I}\right).
\label{eq:residual_append}
\end{align}
\iffalse
We assume that $\delta_i$ and $\left(
\frac{\partial f(\mathrm{LN}(\bm{x)})}{\partial \bm{x}}
\right)_{i,j}$ are independent and the average of $\frac{\partial f(\mathrm{LN}(\bm{x)})}{\partial \bm{x}}$ is zero.
Let $s$ be $E\left[\left\|\frac{\partial f(\mathrm{LN}(\bm{x)})}{\partial \bm{x}}\right\|^2\right]$.
Here, the expectation of the norm of Equation~\ref{eq:residual_append} is
\begin{align*}
&E\left[
\left\|
\bm{\delta} \left(\frac{\partial f(\mathrm{LN}(\bm{x)})}{\partial \bm{x}} +\bm{I}\right)
\right\|^2
\right] \\
&=E\left[
\left\|
\bm{\delta}\frac{\partial f(\mathrm{LN}(\bm{x)})}{\partial \bm{x}}
\right\|^2
\right]
+ E\left[
\left\|
\bm{\delta}
\right\|^2
\right] \\
&=d_{\mathrm{in}} \mathrm{Var}\left[\left\|\frac{\partial f(\mathrm{LN}(\bm{x)})}{\partial \bm{x}} 
\right\|^2\right]E[\|\bm{\delta}\|^2] 
+
E\left[
\|
\bm{\delta}
\|^2
\right] \\
&= (s^2 +1) E\left[
\|
\bm{\delta}
\|^2
\right].
\end{align*}
Thus, a residual connection causes an $(s^2+1)$-fold increase in the squared norm of the gradient $E\left[\left\|\frac{\partial \mathcal{L}}{\partial \bm{x}}\right\|^2\right]$.
\fi
We assume that $\delta_i$ and $
\frac{\partial f(\mathrm{LN}(\bm{x)})_i}{\partial x_j}
$ are independent and the average of $\frac{\partial f(\mathrm{LN}(\bm{x)})_i}{\partial x_j}$ is zero.
Let $s^2$ be $E\left[\left\|\frac{\partial f(\mathrm{LN}(\bm{x)})_i}{\partial \bm{x}}\right\|^2\right]$.
Here, the expectation of the norm of Equation~\ref{eq:residual_append} is
\begin{align*}
E&\left[
\left\|
\bm{\delta} \left(\frac{\partial f(\mathrm{LN}(\bm{x)})}{\partial \bm{x}} +\bm{I}\right)
\right\|^2
\right] \\
&=E\left[
\left\|
\bm{\delta}\frac{\partial f(\mathrm{LN}(\bm{x)})}{\partial \bm{x}}
\right\|^2
\right]
+ E\left[
\left\|
\bm{\delta}
\right\|^2
\right] \\
%&=\mathrm{Var}\left[
%\left\|
%\bm{\delta}\frac{\partial f(\mathrm{LN}(\bm{x)})}{\partial \bm{x}}
%\right\|
%\right]
%+ E\left[
%\left\|
%\bm{\delta}
%\right\|^2
%\right] \\
%&=d_\mathrm{in}\mathrm{Var}\left[
%\left\|
%\bm{\delta}\frac{\partial f(\mathrm{LN}(\bm{x)})}{\partial x_j}
%\right\|
%\right]
%+ E\left[
%\left\|
%\bm{\delta}
%\right\|^2
%\right] \\
%&=d_\mathrm{in} d_\mathrm{out} \mathrm{Var}\left[
%\left\|
%\delta_i \frac{\partial f(\mathrm{LN}(\bm{x)})_i}{\partial x_j}
%\right\|
%\right] \\
%& \quad + d_\mathrm{out} E\left[
%\left\|
%\delta_i
%\right\|^2
%\right] \\
&=d_\mathrm{in} d_\mathrm{out} E\left[
\left\|
\delta_i 
\right\|^2
\right]
E\left[
\left\|
\frac{\partial f(\mathrm{LN}(\bm{x)})_i}{\partial x_j}
\right\|^2
\right] \\
& \quad + d_\mathrm{out} E\left[
\left\|
\delta_i
\right\|^2
\right] \\
&= \left(d_\mathrm{in} E\left[
\left\|
\frac{\partial f(\mathrm{LN}(\bm{x)})_i}{\partial x_j}
\right\|^2
\right] +1\right) 
E\left[
\|
\bm{\delta}
\|^2
\right] \\
&= (s^2 +1) E\left[
\|
\bm{\delta}
\|^2
\right].
\end{align*}
Thus, a residual connection causes an $(s^2+1)$-fold increase in the squared norm of the gradient $E\left[\left\|\frac{\partial \mathcal{L}}{\partial \bm{x}}\right\|^2\right]$.

\iffalse
&=d_{\mathrm{in}} E\left[
\left\|
\bm{\delta}\frac{\partial f(\mathrm{LN}(\bm{x)})}{\partial x_j}
\right\|^2
\right]
+
E\left[
\|
\bm{\delta}
\|^2
\right] \\
&=d_{\mathrm{in}} E \left[\left\|\frac{\partial f(\mathrm{LN}(\bm{x)})}{\partial x_j} 
\right\|^2\right]E[\|\bm{\delta}\|^2] 
+
E\left[
\|
\bm{\delta}
\|^2
\right] \\
&= (s^2 +1) E\left[
\|
\bm{\delta}
\|^2
\right].
\fi

\section{Experimental Setup}
\label{append:experimental_setup}

\begin{table}[t!]
\centering
    %\scalebox{0.75}{
    \small
    \tabcolsep3pt
		\begin{tabular}{l|ccc} \toprule
                & 130M & 1.3B & 13B \\
                \midrule
                Hidden Size $d$ & 768 & 2048 & 5120 \\   
                \# Layer $N$ & 12 & 24 & 40 \\   
                \# Attention Head & 12 & 16 & 40 \\ 
                Context Length & \multicolumn{3}{c}{2048} \\ 
                Vocabulary Size & \multicolumn{3}{c}{32000} \\
                RMSNorm $\epsilon$ & \multicolumn{3}{c}{1e-5} \\
                Positional Encoding & \multicolumn{3}{c}{RoPE} \\
                Bias in Linear & \multicolumn{3}{c}{none} \\
                \bottomrule
        \end{tabular}%}
	\caption{Detailed model configuration.} \label{tab:model_config_detail}
\end{table}

\begin{table}[t!]
\centering
    %\scalebox{0.75}{
    \small
    \tabcolsep3pt
		\begin{tabular}{l|cc} \toprule
                & Rapid Setting & Stable Setting \\ \midrule
                Batch Size [tokens] & 1M & 4M \\
                Learning rate $\mu$ & 1e-3 & 5e-4 \\
                Warmup Steps & 100 & 2000 \\
                Precision & \multicolumn{2}{c}{bfloat16} \\
                Corpus Size [tokens] & \multicolumn{2}{c}{30B} \\
                Adam $\beta_1$ & \multicolumn{2}{c}{0.9} \\
                Adam $\beta_2$ & \multicolumn{2}{c}{0.95} \\
                Gradient Clipping Threshold & \multicolumn{2}{c}{1} \\
                Weight decay & \multicolumn{2}{c}{0.01} \\  
                Z-loss & \multicolumn{2}{c}{1e-4} \\ \bottomrule
        \end{tabular}%}
	\caption{Detailed training configuration.}
 \label{tab:optim_config_detail}
\end{table}

Table~\ref{tab:model_config_detail} and \ref{tab:optim_config_detail} list the detailed model and training configurations, respectively.
We used eight NVIDIA H100 (80GB) GPUs for pre-training the 130M and 1.3B models and 64 GPUs for pre-training the 13B models.
The pre-trainings took roughly 12 hours, 60 hours, and 40 hours, respectively. 
We used the Adam optimizer~\cite{adam}, PyTorch (ver.~2.1.0)\footnote{\url{https://pytorch.org/}}~\cite{paszke2017automatic}, transformers (ver.~4.37.2)\footnote{\url{https://github.com/huggingface/transformers}}~\cite{wolf2019huggingface}, and llm-foundry (ver.~0.5.0)~\footnote{\url{https://github.com/mosaicml/llm-foundry}}.

\section{Relation to Existing Reparameterization Methods}
\label{append:reparam_compare}
\subsection{Weight Normalization}
Weight Normalization~\cite{NIPS2016_ed265bc9} conducts L2 normalization and scaling of each row of the parameter matrix $\bm{w} \in \mathbb{R}^{d_\mathrm{in}}$:
\[
\bar{\bm{w}} = \frac{\alpha}{\|\bm{w}\|}\bm{w}.
\]
It differentiates the whole operation including the normalization and propagates the gradient to $\bm{w}$. It determines the initial $\alpha$ from the value of the forward computation in the first step. The proposed method is efficient because it does not conduct normalization and provides a matrix-wise reparameterization; the number of the additional parameter $\alpha$ per parameter matrix is one. % in the proposed method and $d_\mathrm{out}$, which is the number of rows in the matrix, in Weight Normalization.

%つまり，Weight Normalizationはパラメータ行列あたり$d_{\mathrm{out}}$のパラメータが追加で必要である．LLMではbias項すら用いずにパラメータを節約することが一般的であり，$d_{\mathrm{out}}$個の追加の影響は大きい．
%また，Weight Normalizationの$\alpha$の初期値は1であり，提案手法のように初期値の制約をパラメータから除外する意図を持った手法ではない．

\subsection{ \texorpdfstring{$\mathrm{\sigma}$}{σ}Reparam}
$\mathrm{\sigma}$Reparam~\cite{pmlr-v202-zhai23a} conducts spectral normalization and scaling of the parameter matrix $\bm{W} \in \mathrm{R}^{d_\mathrm{out} \times d_\mathrm{in}}$,
\[
\bar{\bm{W}} = \frac{\alpha}{\|\bm{W}\|_2}\bm{W},
\]
where $\|\bm{W}\|_2$ is the spectral norm (\textit{i.e.,} the maximum singular value).
The maximum singular value is calculated by the power method that is iterated once per batch~\cite{miyato2018spectral}. It does not differentiate the spectral normalization.
$\mathrm{\sigma}$Reparam is based on the fact that the entropy in the self-attention affects the stability of the training. It regulates the singular value of $\bar{\bm{W}}$ so as to control the entropy. $\alpha$ is initialized to 1.
Therefore, $\mathrm{\sigma}$Reparam is different from the proposed method, which is designed to align the virtual parameter $\alpha_{\cdot} \bm{W}_{\cdot}$ to any initialization algorithm, such as He initialization, while setting the standard deviations of the actual parameter $\bm{W}_{\cdot}$ independently.

\subsection{Residual Scaling as Reparameterization}
\citet{NEURIPS2022_ae0cba71} overcomes the limitation of the $(1/\sqrt{2N})$-fold multiplication of $\sigma_o$ and $\sigma_d$ caused by the residual connection (Equation \ref{eq:residual_forward}). It modifies the residual connection to 
\[
\bm{y} =\frac{1}{\sqrt{2N}}f(\mathrm{LN}(\bm{x)}) + \bm{x}.
\]
Different from the original residual scaling, which changes the standard deviations, this equation can be viewed as a reparameterization of $\bm{W}_o$ and $\bm{W}_d$ because of its linearity. The proposed method 
can be interpreted as a generalization of the reparameterization to all parameters. Because of the generalization, the proposed method overcomes any limitations to the norms of the parameters that is caused by an initialization algorithm. Therefore, it can determine a common $\sigma$ for all parameters even without a dependence on $d$. Also, it makes the gate parameters trainable.

\section{Comparison of Reparameterization Methods on 30B Tokens}
\label{append:full_run}
\begin{table}[t!]
\centering
    %\scalebox{0.75}{
    \small
    \tabcolsep3pt
		\begin{tabular}{l|ccccc} \toprule
                & WikiText & LAMBADA \\
                \midrule
                Small Init. & 16.55 & 26.29  \\ \midrule
                Weight Normalization & \textbf{14.13} & 24.97  \\ \midrule
                %WeightNorm. w./ matrix reparam. & 15.05 & 23.68   \\ \midrule
                $\mathrm{\sigma}$Reparam w./ Pre-LN & 18.83 & 26.22   \\ 
                $\mathrm{\sigma}$Reparam w./ Post-LN & 16.52 & 25.58 \\ \midrule
                Residual Scaling w./ He Init.& 19.05 & 27.36  &  \\ 
                Residual Scaling w./ Small Init. & 18.03 & 26.88 \\ \midrule
                WeSaR & 14.51 & \textbf{22.87}  \\
                \bottomrule
        \end{tabular}%}
	\caption{Comparison of reparameterization methods on 30B tokens.} \label{tab:reparam_result_full}
\end{table}
We compared WeSaR with the existing reparameterization methods on 30B tokens. The results shown in Table~\ref{tab:reparam_result_full} achieved the same tendency as the results on 10B tokens. 
In particular, similar to the results of five runs on 10B tokens in Table~\ref{tab:reparam_result_std}, Weight Normalization achieved comparable performance.
However, Weight Normalization took the longest time for the training due to the back-propagation through the normalization. Thus, WeSaR's simple reparameterization is efficient and effective for LLM's pre-training.
%Although we conducted 30B pre-training only once due to the computational resources, we consider that this difference is comparable 
%Here, compared to the matrix-wise reparameterization in WeSaR, the vector-wise one in Weight Normalization reduced the perplexity in WikiText in which the models achieved smaller perplexity, but increased it in LAMBADA in which it did larger perplexity. Therefore, we consider that the vector-wise reparameterization tends to cause overfitting to the trained domain although there is little difference in the number of parameters. We concluded that our matrix-wise reparameterization is both efficient and effective for LLM initialization. 
%Also, from the comparison of the proposed method and the matrix-wise Weight Normalization, we consider that scaling parameters without normalization is sufficient for the LLM pre-training.

\section{Loss Values without Loss Spikes}
\label{append:loss_decrease}
\begin{figure}[t!]
\centering
		\includegraphics[width=0.8\linewidth]{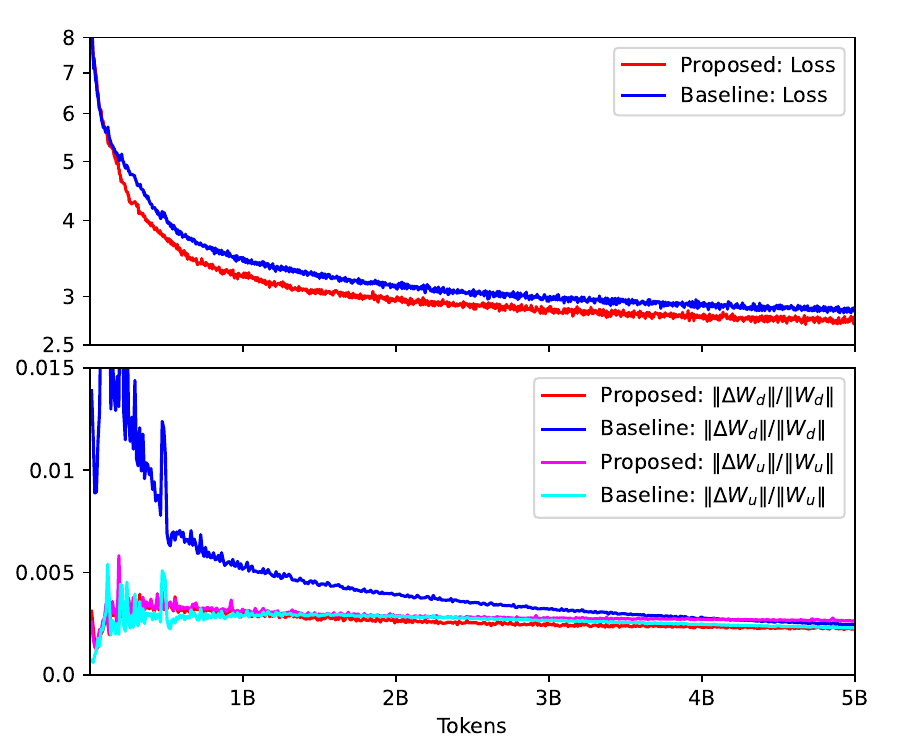}
		\caption{Loss of the 1.3B Transformer models at the beginning of the training (top). Update ratios $\|\Delta \bm{W}_d\|/\| \bm{W}_d\|$ and $\|\Delta \bm{W}_u\|/\| \bm{W}_u\|$ of the same (bottom).}
        \label{fig:1.3B_init_loss}
\end{figure}

\begin{figure}[t!]
\centering
		\includegraphics[width=0.8\linewidth]{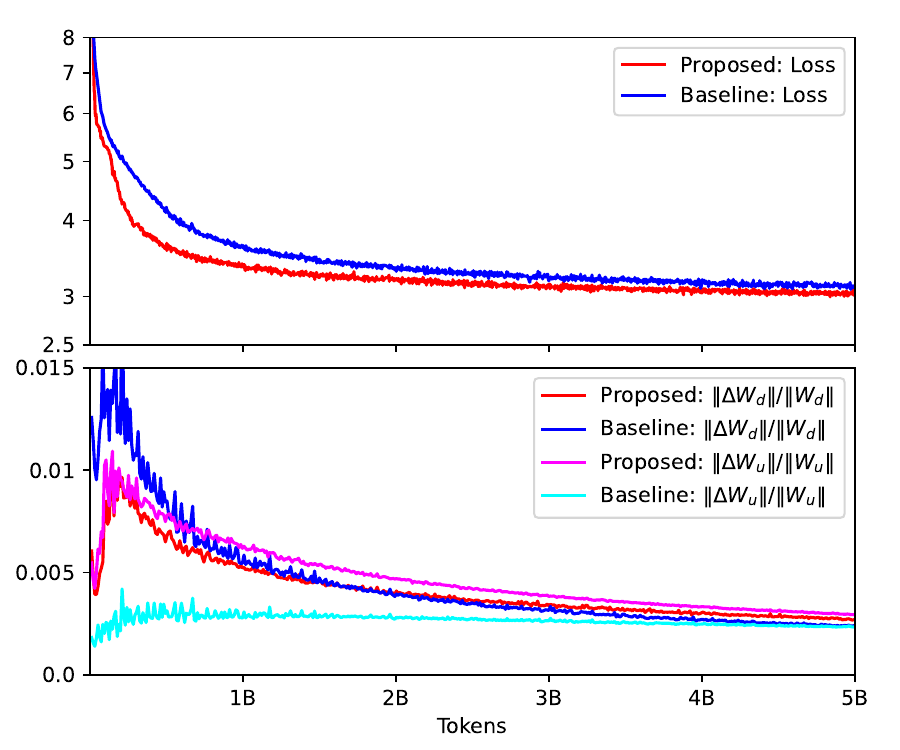}
		\caption{Loss of the 130M Transformer models at the beginning of the training (top). Update ratios $\|\Delta \bm{W}_d\|/\| \bm{W}_d\|$ and $\|\Delta \bm{W}_u\|/\| \bm{W}_u\|$ of the same (bottom).}
        \label{fig:130M_init_loss}
\end{figure}
\begin{figure}[t!]
\centering
		\includegraphics[width=0.8\linewidth]{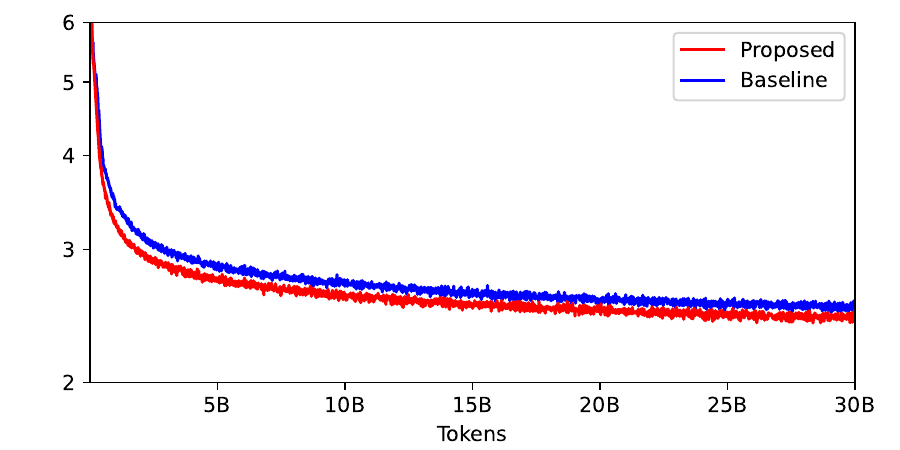}
		\caption{Loss of 1.3B models during training.}
        \label{fig:1.3B_loss}
\end{figure}
\begin{figure}[t!]
\centering
		\includegraphics[width=0.8\linewidth]{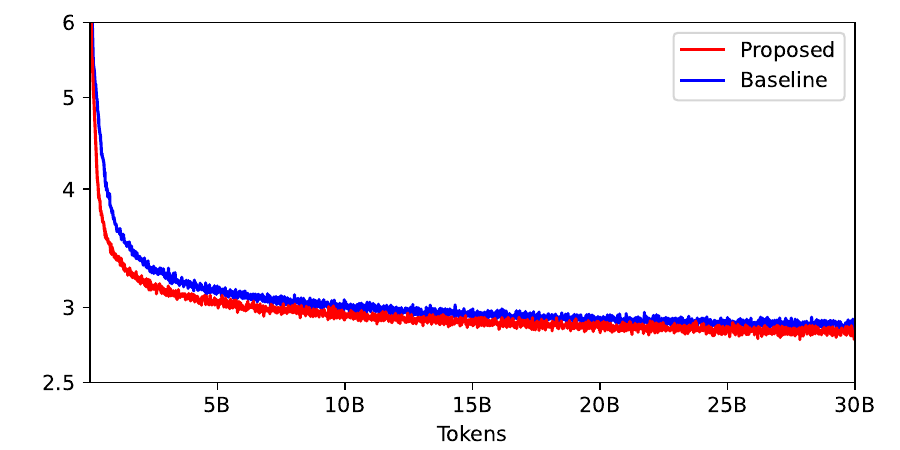}
		\caption{Loss of 130M models during training.}
        \label{fig:130M_loss}
\end{figure}

Figure~\ref{fig:1.3B_init_loss} and \ref{fig:130M_init_loss} show the loss decrease and the update ratios at the beginning of the training of the 1.3B models and the 130M models, respectively.
Because the 1.3B and 130M models did not cause loss spikes, we did not observe a drastic decrease in the update ratios like with the 13B models. 
Except for that point, the update ratio behaved similarly to the 13B models. We should note that, in smaller models, the effect of $(1/\sqrt{2N})$-fold scaling gets smaller, and thus there is less difference between the baseline method and WeSaR.
Figure~\ref{fig:1.3B_loss} and \ref{fig:130M_loss} show the loss values during the training.

Moreover, we confirmed that WeSaR outperformed Small initialization both in the loss values in Figure~\ref{fig:13B_init_loss},~\ref{fig:13B_loss},~\ref{fig:1.3B_init_loss},~\ref{fig:130M_init_loss},~\ref{fig:1.3B_loss}, and \ref{fig:130M_loss} and the perplexity in Table~\ref{tab:main_result}. We consider that the small standard deviation $\sigma^2 =\textrm{4e-5}$, which corresponds to $d =10,000$ in the Small initialization criteria, accelerated the training.

\begin{figure}[t!]
\centering
		\includegraphics[width=1.\linewidth]{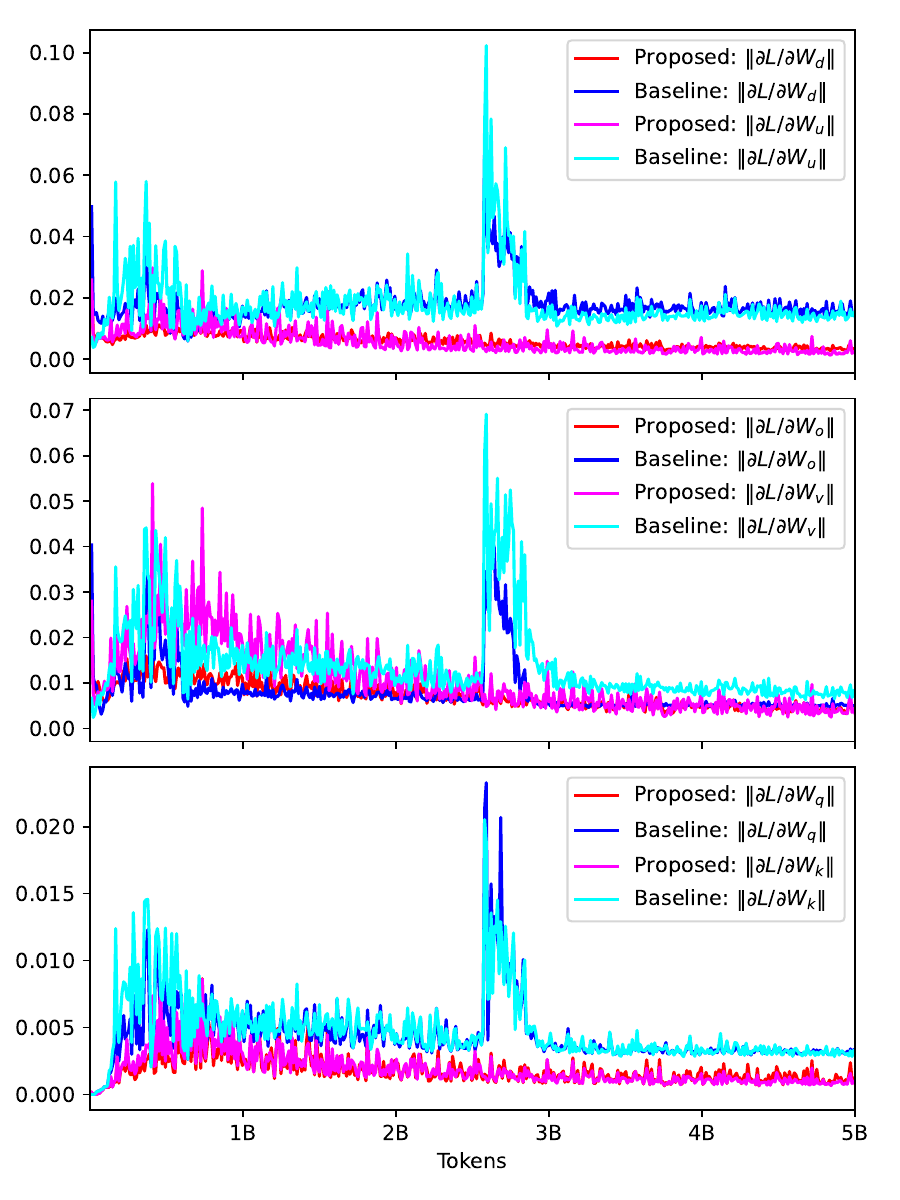}
		\caption{Norm of the gradient of the parameters at the 40th layer of 13B models during training.}
        \label{fig:13B_39_update}
\end{figure}

\section{Update Ratios in Other Layers and Comparison with Gradient Norm}
\label{append:relative_update}
\begin{figure}[t!]
\centering
		\includegraphics[width=1.\linewidth]{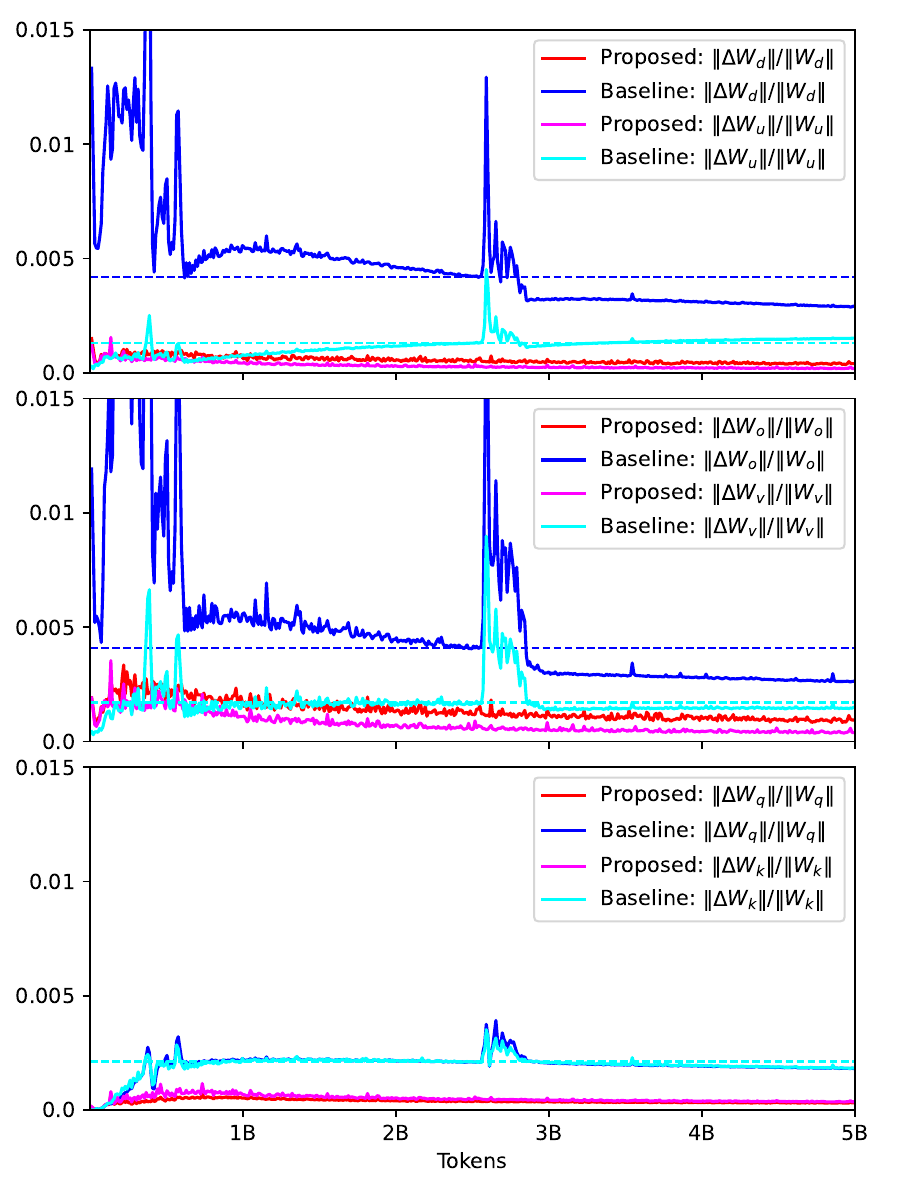}
		\caption{Update ratio at the 40th layer of 13B models during training.}
        \label{fig:13B_39_rel}
\end{figure}

\begin{figure}[t]
\centering
		\includegraphics[width=1.\linewidth]{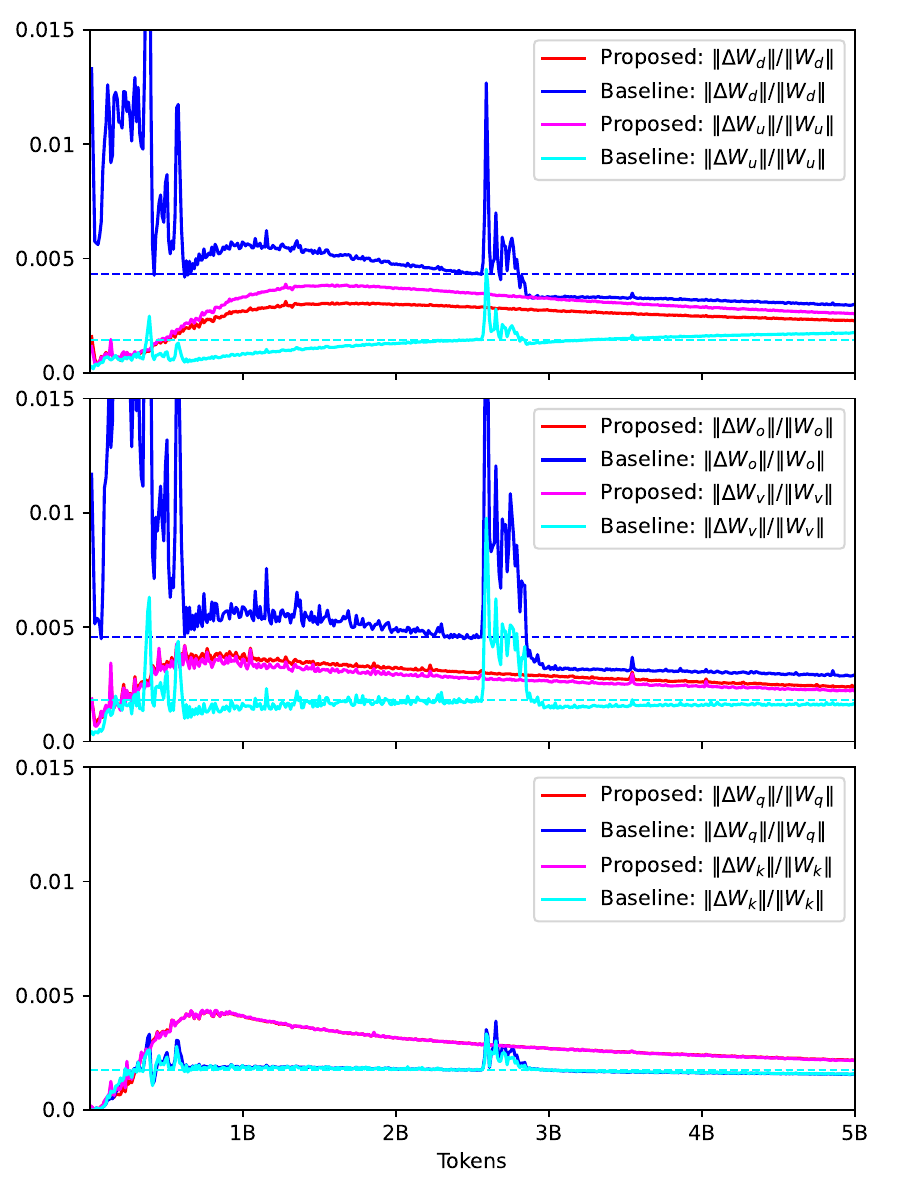}
		\caption{Update ratio at the 27th layer of 13B models during training.}
        \label{fig:13B_26_rel}
\end{figure}

\begin{figure}[t]
\centering
		\includegraphics[width=1.\linewidth]{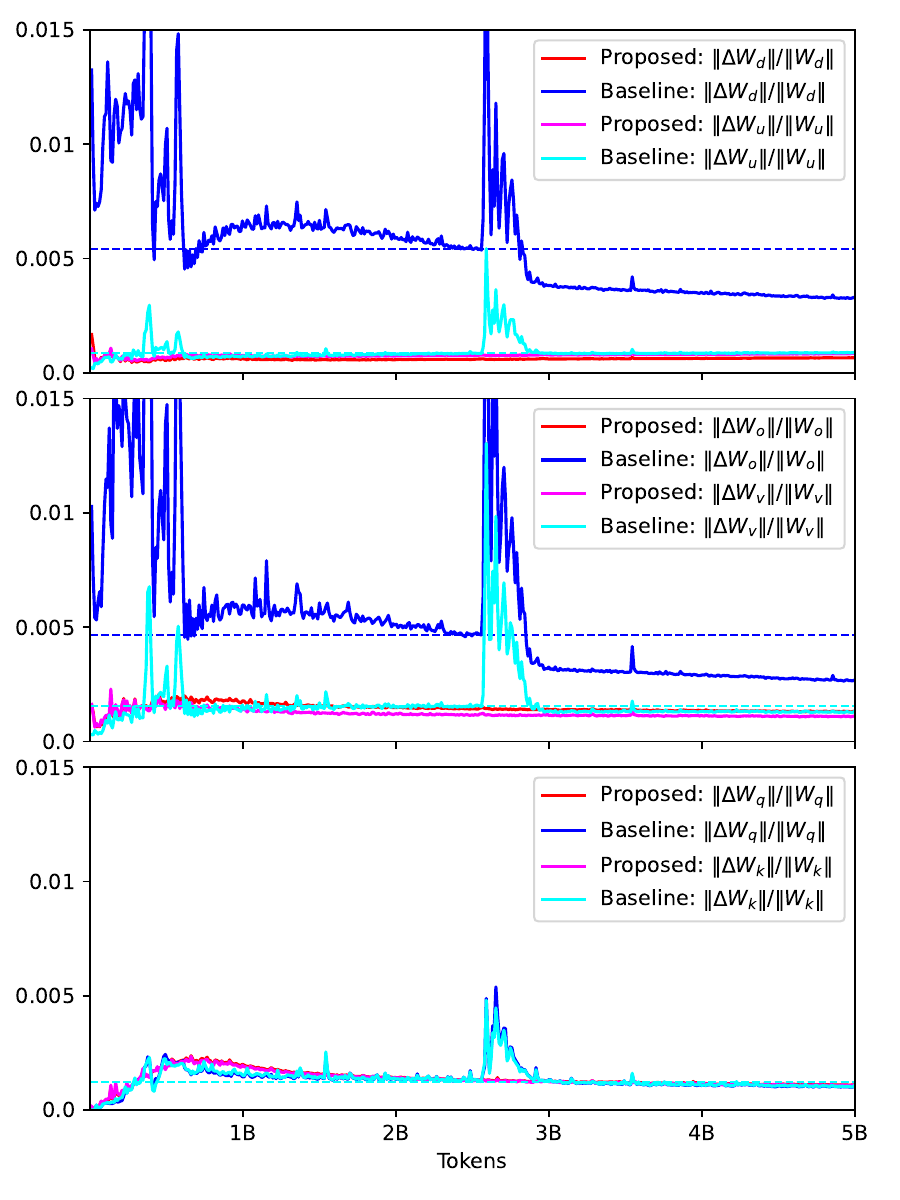}
		\caption{Update ratio at the 14th layer of 13B models during training.}
        \label{fig:13B_13_rel}
\end{figure}

\begin{figure}[t]
\centering
		\includegraphics[width=1.\linewidth]{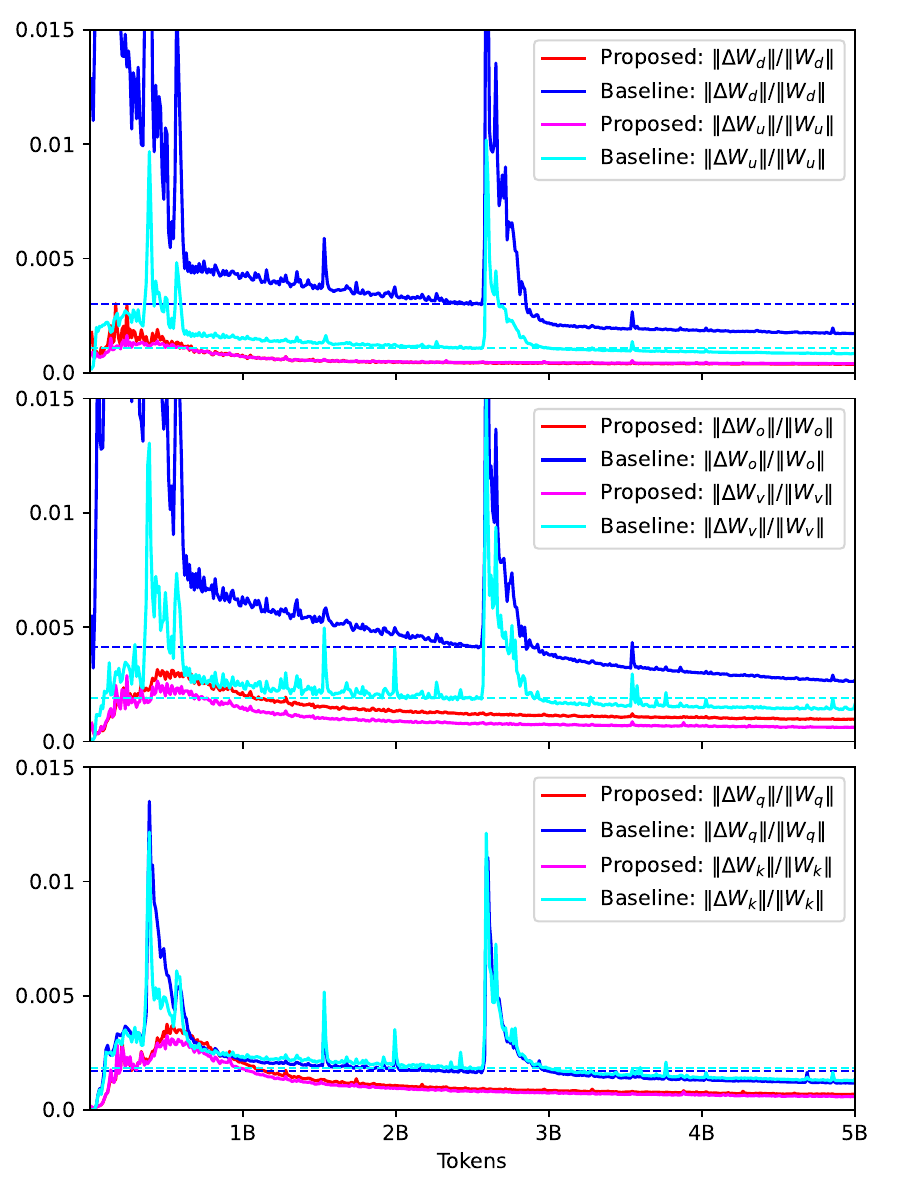}
		\caption{Update ratio at the 1st layer of 13B models during training.}
        \label{fig:13B_0_rel}
\end{figure}

Figure~\ref{fig:13B_39_rel}, \ref{fig:13B_26_rel}, \ref{fig:13B_13_rel}, and \ref{fig:13B_0_rel} show the update ratios $\|\Delta \bm{W}_{\cdot} \|/ \|\bm{W}_{\cdot} \|$ of all linear layers at the 40th, 27th, 14th, and 1st Transformer layers in the 13B models, respectively. Because the 1.3B and 130M models did not cause loss spikes as shown in Figure~\ref{fig:1.3B_init_loss} and Figure~\ref{fig:130M_init_loss}, we only list the update ratios of the 13B models. 

We observed that the update ratios $\|\Delta \bm{W}_d \|/ \|\bm{W}_d \|$ and $\|\Delta \bm{W}_o \|/ \|\bm{W}_o \|$ 
in the baseline method decreased after loss spikes, except for $\bm{W}_o$ in the 1st layer. We consider that $\bm{W}_d$ and $\bm{W}_o$, the parameters whose norm is smaller than the others, caused loss spikes due to their large update ratios. We also confirmed that the update ratios trained with WeSaR were stable among all layers and all parameters.

The existing studies that tackled loss spikes~\cite{glm,pmlr-v202-zhai23a,molybog2023theory,takase2023spike} focused on the gradient norm $\|\frac{\partial \mathcal{L}}{\partial \bm{W}}\|$ as a clue to understand loss spikes. However, instead of the gradient norm itself, we focused the update ratio. Figure~\ref{fig:13B_39_update} shows the norm of the gradients of the parameters at the last layer of the 13B models, which corresponds to the update ratios shown in Figure~\ref{fig:13B_39_rel}.
We observed that a phenomenon of a drastic change in scale before and after loss spikes only appeared in the update ratio. Thus, we introduced the update ratio as a novel clue to understand loss spikes.
%Although the scale of the gradient norm before and after the loss spike did not change especially in $\bm{W}_o$, the update ratios ware faced with the drastic decrease at the loss spike.

\clearpage

\end{document}